%% file: Constrained-BRT-VIS2019.tex
\documentclass[10pt,journal,compsoc]{IEEEtran} % add from vica
\usepackage[nocompress]{cite} % add from vica
\usepackage[bookmarks,backref=true,linkcolor=black]{hyperref} %,colorlinks add from vica
\usepackage{verbatim} % add from vica
\usepackage{ulem} 
\usepackage{makecell}
\usepackage{mathptmx}
\usepackage[pdftex]{graphicx}
\usepackage{epstopdf}
\usepackage{times}
\usepackage[ruled,vlined]{algorithm2e}  %change from vica  
\usepackage{amssymb,amsmath}
\usepackage{multirow}
\usepackage{subfigure}
\usepackage{paralist}
\usepackage{tikz}
\usepackage{multirow}
\usepackage{color}
\usepackage{ulem}
\usepackage{setspace}
\usepackage{microtype,hyphenat,balance}
\usepackage{array}
\usepackage{url}
\usepackage{bm}

\usepackage{amsfonts}
\usepackage{bbm}
\usepackage{diagbox}
\usepackage{url}
\usepackage{caption}
\usepackage{wrapfig}

\usepackage{ragged2e}

\everymath{\color{black}}

\definecolor{mypink}{RGB}{255, 100, 130}
\definecolor{mygreen}{RGB}{106, 168, 79}
\newcommand{\xiting}[1]{\textcolor{black}{#1}}
\newcommand{\xitingrevision}[1]{\textcolor{black}{#1}}
\newcommand{\wenwen}[1]{\textcolor{black}{#1}}
\newcommand{\shixia}[1]{\textcolor{black}{#1}}

\newcommand{\weikai}[1]{\textcolor{black}{#1}}
\newcommand{\weikaiminor}[1]{\textcolor{black}{#1}}

\let\oldtextbf\textbf
\renewcommand{\textbf}[1]{\oldtextbf{\normalsize{#1}}}
\let\oldtextit\textit
\renewcommand{\textit}[1]{\oldtextit{\normalsize{#1}}}

\begin{document}
%\setlength{\parskip}{0.1pt}
%\setlength{\parskip}{3.1pt}

% \section{Related Work}
% \begin{itemize}
% 	\item Uncertainty Visualization
% 	\item Uncertainty Modeling
% 	\item Multivariate Visualization
% \end{itemize}

% \title{Interactive Steering of Constrained Hierarchical Clustering}
\title{Interactive Steering of Hierarchical Clustering}

\author{
		Weikai~Yang, Xiting~Wang, Jie~Lu, Wenwen~Dou, Shixia~Liu
\IEEEcompsocitemizethanks{
\IEEEcompsocthanksitem W. Yang, J. Lu, and S. Liu are with Tsinghua University. %\protect\\
% E-mail: {shixia}@tsinghua.edu.cn.
\IEEEcompsocthanksitem X. Wang is with Microsoft Research. 
% E-mail: xitwan@microsoft.com
\IEEEcompsocthanksitem W. Dou is with University of North Carolina at Charlotte
% \IEEEcompsocthanksitem B. Guo is with Microsoft Research.  %\protect\\
% E-mail: bainguo@microsoft.com.
}% <-this % stops a space
\thanks{}}
%

%\markboth{IEEE Transactions on Visualization and Computer Graphics,~Vol.~6, No.~1, January~2007}%
%{Chen \MakeLowercase{\textit{et al.}}: TextPioneer: Exploring Topical Lead-Lag Evolution across Corpora}

\IEEEcompsoctitleabstractindextext{%
\justify
\begin{abstract}
    %Hierarchical clustering is an important part of exploratory data analysis, especially for big data. 
    Hierarchical clustering is an important technique to organize big data for exploratory data analysis.
    However, existing one-size-fits-all hierarchical clustering methods often fail to meet the diverse needs of different users.
    To address this challenge, we present an interactive steering method to visually supervise constrained hierarchical clustering by utilizing both public knowledge (e.g., Wikipedia) and private knowledge from users.
    % from the available ontology repositories (e.g., Wikipedia) and private knowledge from users. 
    The novelty of our approach includes 
    % 1) automatically constructing constraints for \weikai{otherwise?} unsupervised hierarchical clustering using knowledge, 
    1) automatically constructing constraints for hierarchical clustering using knowledge (knowledge-driven) and intrinsic data distribution (data-driven), 
    and 2) enabling the interactive steering of clustering through a visual interface (user-driven). 
    % To better leverage public knowledge, 
    Our method first maps each data item to the most relevant items in a knowledge base.
    An initial constraint tree is then extracted using the ant colony optimization algorithm. 
    The algorithm balances the tree width and depth and covers the data items with high confidence.
    %by combining a pruned breadth-first search with the ant colony optimization algorithm. 
    %Such a combination well balances the tree width and depth, and covers more documents with high confidence.
    Given the constraint tree, the data items are hierarchically clustered using evolutionary Bayesian rose tree. 
    To clearly convey the hierarchical clustering results, an uncertainty-aware
    %uncertainty-aware
    tree visualization has been developed to 
    % The tree visualization as well as uncertainty contributed by the model, public knowledge, and domain knowledge 
     enable users to quickly locate the most uncertain sub-hierarchies and interactively improve them. 
    The quantitative evaluation and case study demonstrate that the proposed approach facilitates the building of customized clustering trees in an efficient and effective manner.
    % , without requiring laborious input from end users. % less human efforts. 
\end{abstract}
% Note that keywords are not normally used for peer review papers.
}
\teaser{
\setcounter{figure}{0}
\centering
\vspace{-14pt}
  \includegraphics[trim=1 1 1 1, clip,width=0.9\linewidth]{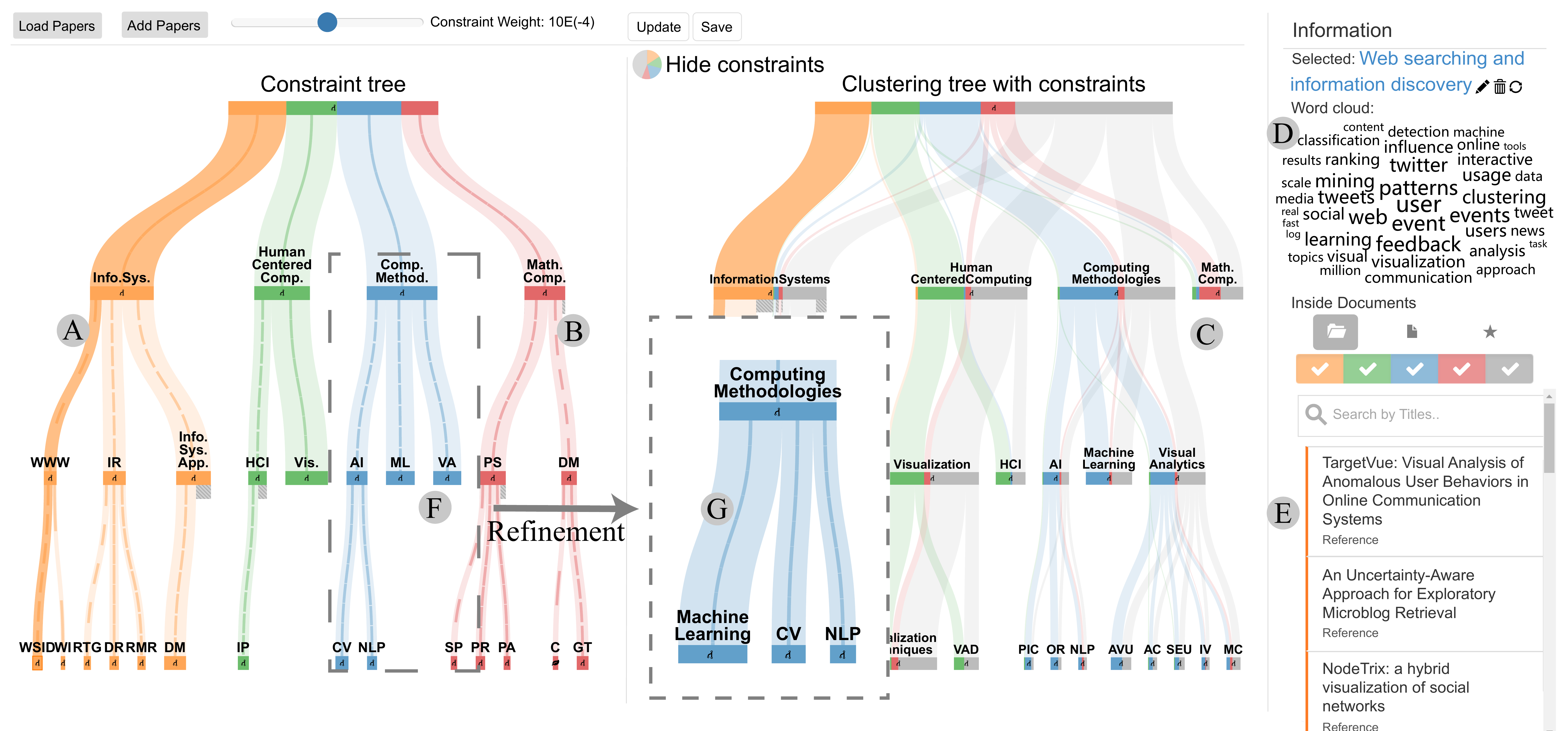}
  \put(-475,210){\small(a)}
  \put(-475,190){\small(b)}
  \put(-110,190){\small(c)}
  \put( -20,210){\small(d)}
  \vspace{-1mm}
  \captionsetup{width=0.91\linewidth}
  \caption{
    ReVision: 
    % (a) the control panels to load the constraints and new documents, adjust the constraint weight, recluster the documents, and save the results; 
    (a) the control panel to load constraints and update clustering results;
    (b) the constraint tree;
    % (c) the tree view for the constraint and the clustering result, user can drag the clusters to adjust the structure; 
    % (c) the hierarchical clustering results. The colors encode the first level clusters of the constraint tree;
    (c) the hierarchical clustering results. The colors encode the first-level categories of the constraint tree;
    (d) the information panel to facilitate understanding and customization of clustering.
    % for the most recently clicked node, users can gain more detail for clusters and the corresponding documents, and add/remove the documents for the cluster. 
    }
\vspace{-6mm}
\label{fig:teaser-overview}
}

\maketitle
\IEEEdisplaynotcompsoctitleabstractindextext
\IEEEpeerreviewmaketitle

{
\fontsize{10}{10} %this command adjusts the fontsize of all the mathmatical symbols in the paper
\input{introduction.tex}

\input{related.tex}

\input{motivation-design-system.tex}
\input{data-modeling.tex}
\input{visualization.tex}
\input{application.tex}

\input{discussion.tex}

\input{conclusion.tex}
}
% begin from vica
% \IEEEdisplaynotcompsoctitleabstractindextext

% \fi
% \IEEEpeerreviewmaketitle

% or


\ifCLASSOPTIONcompsoc
%\else
%\fi

\ifCLASSOPTIONcaptionsoff
  \newpage
\fi



%\bibliographystyle{IEEEtran}
\bibliographystyle{IEEEtran}
\bibliography{reference}
\vspace{-12mm}
% end from vica

%% if specified like this the section will be ommitted in review mode
% \looseness=-1}
% \newpage
% \clearpage

%\bibliographystyle{abbrv} % disable from vica
%\addtolength{\itemsep}{5ex} % disable from vica
%\bibliography{reference}

  % sigproc.bib is the name of the Bibliography in this case

%1 page


% \fi
\begin{IEEEbiography}
[{\includegraphics[width=1in,height=1.25in,clip,keepaspectratio]{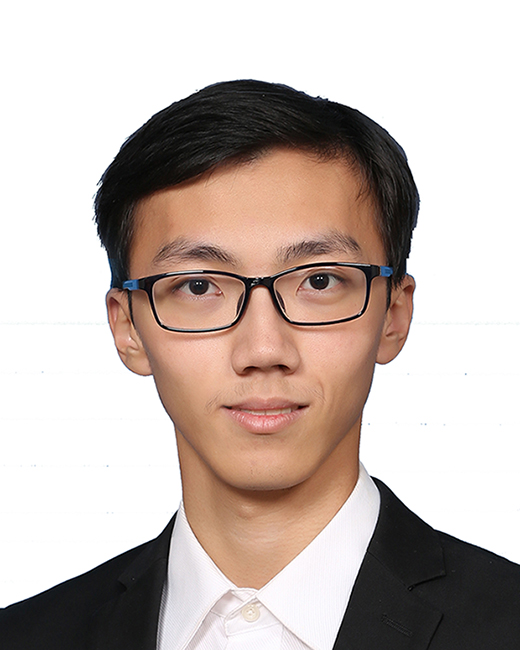}}]{Weikai Yang}
is a graduate student at Tsinghua University. His research interest is
visual text analytics. He received a B.S. degree from Tsinghua University.
\end{IEEEbiography}
\begin{IEEEbiography}
[{\includegraphics[width=1in,height=1.25in,clip,keepaspectratio]{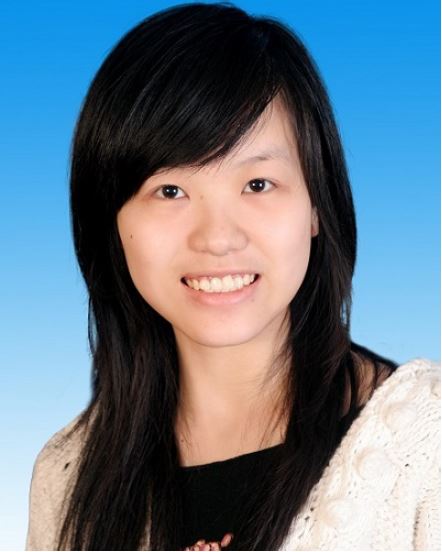}}]{Xiting Wang}
is a senior researcher at Microsoft Research Asia. Her research interests include visual text analytics, text mining, and explainable AI.  She received her Ph.D. degree in Computer Science from Tsinghua University and a B.S. degree in Electronics Engineering from Tsinghua University.
\end{IEEEbiography}
\begin{IEEEbiography}
[{\includegraphics[width=1in,height=1.25in,clip,keepaspectratio]{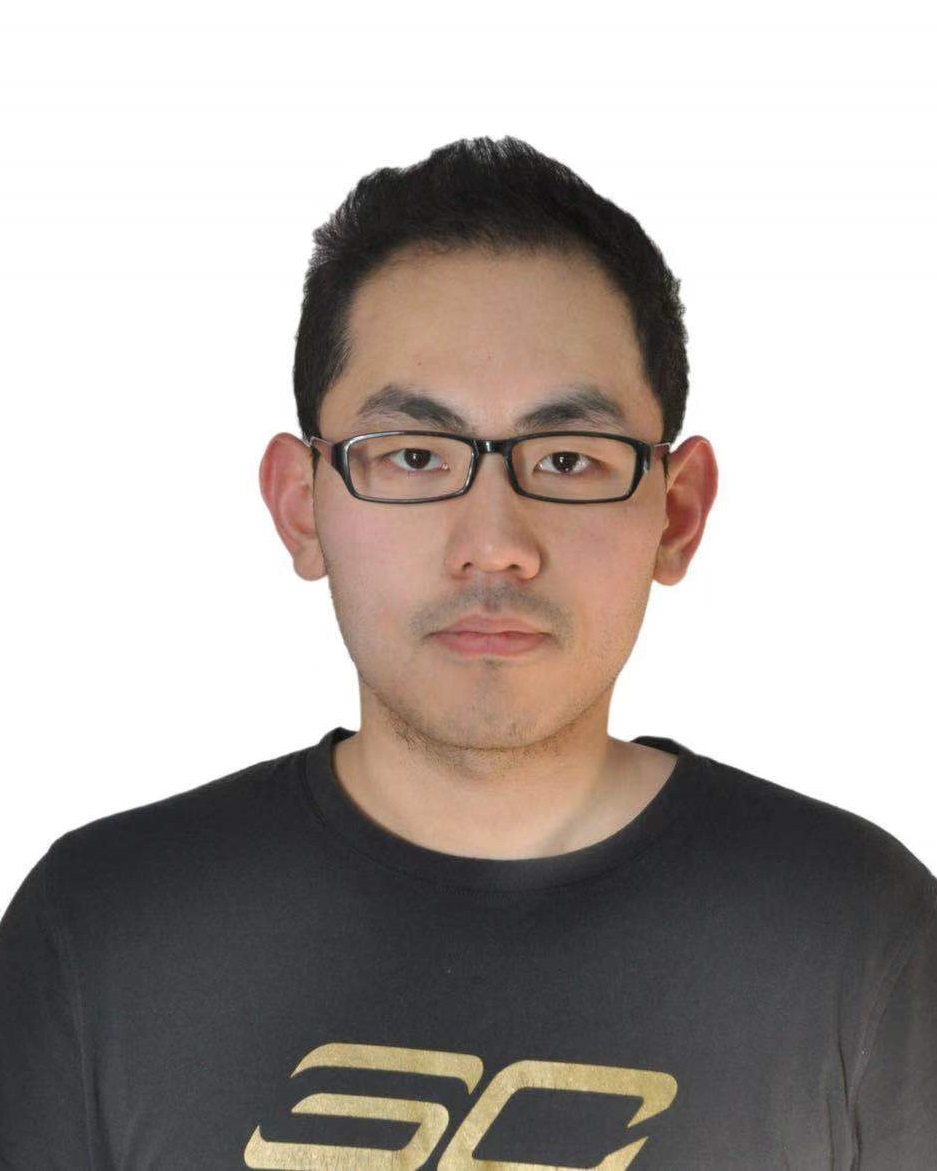}}]{Jie Lu}
is a software engineer at Microsoft STCA. His research interests include graph visualization and visual text analytics. He received a master degree from Tsinghua University and B.S. degree from Tongji University.
\end{IEEEbiography}
\begin{IEEEbiography}
[{\includegraphics[width=1in,height=1.25in,keepaspectratio]{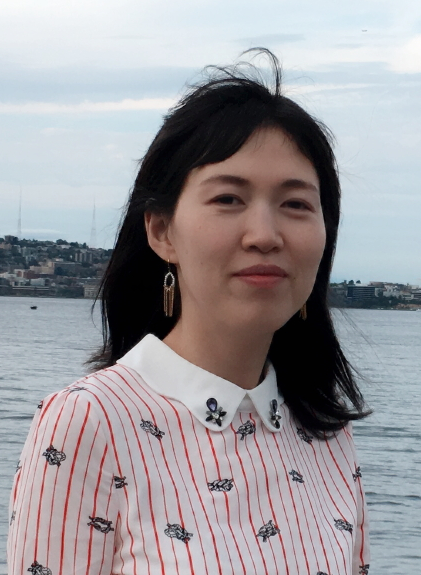}}]{Wenwen Dou} is an assistant professor  at  the  University  of  North  Carolina  at Charlotte. Her research interests include Visual Analytics, Text Mining, and Human Computer Interaction. Dou has worked with various analytics domains in reducing information overload and providing interactive visual means to analyzing unstructured information. She has experience in turning cutting-edge research into technologies that have broad societal impacts.
\end{IEEEbiography}

\begin{IEEEbiography}
[{\includegraphics[width=1in,height=1.25in,clip,keepaspectratio]{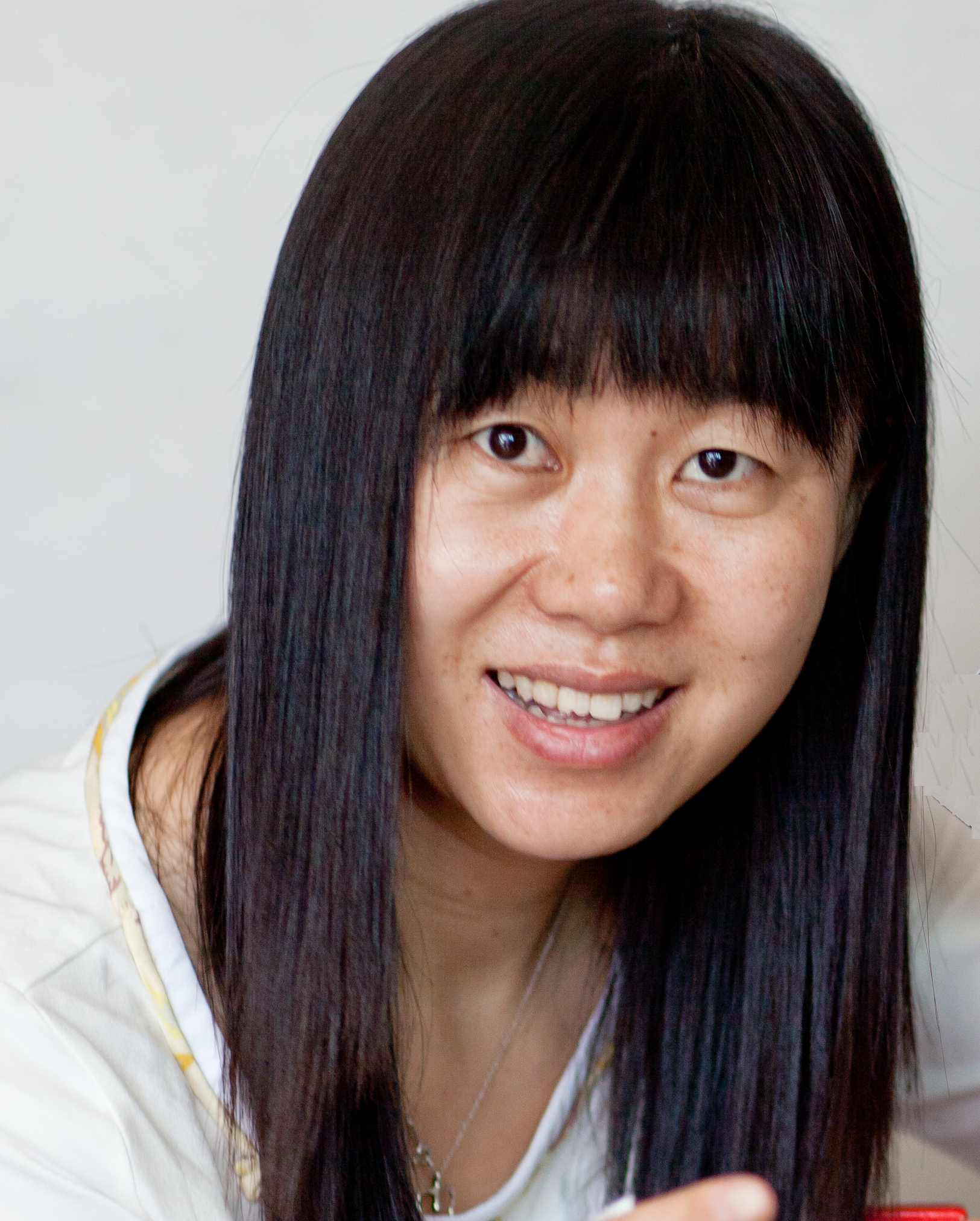}}]{Shixia Liu}
is an associate professor at Tsinghua University. Her research interests include visual text analytics, visual social analytics, interactive machine learning, and text mining. She worked as a research staff member at IBM China Research Lab and a lead researcher at Microsoft Research Asia.
She received a B.S. and M.S. from Harbin Institute of Technology, a Ph.D. from Tsinghua University.
She is an associate editor-in-chief of IEEE Trans. Vis. Comput. Graph.
\end{IEEEbiography}
\end{document}

%% file: introduction.tex
\section{Introduction} 
\maketitle 

Rich hierarchies are ubiquitous in data sets across many disciplines, including topic hierarchies in text corpora, hierarchical communities in social media, and the structured organization of image databases~\cite{blei2010nested,blundell2010bayesian,liang2018photorecomposer,liu2017towards,liu2018analyzing,wang2016topicpanorama}.
The capability of hierarchies to illustrate relationships among data instances and summarize data at different granularities makes them very useful in exploratory data analysis~\cite{liu2018bridging}.

Despite the aforementioned benefits of hierarchies, there is a lack of a systematic process for constructing effective hierarchies based on individual user needs.
Practitioners usually struggle with the construction of hierarchical clusters due to the unsupervised nature of existing algorithms and the cognitive complexity of manual solutions.
On the one hand, data is often clustered in ways that do not suit diverse user needs \cite{Kwon2018Clustervision,Cavallo2019Clustrophile}.
Moreover, the absence of labels for the validation of clusters and hierarchies prevents existing unsupervised algorithms from producing satisfying task-relevant results.
On the other hand, the possible number of candidate hierarchies is super-exponential to the number of data instances~\cite{blundell2010bayesian}.
As a result, manually building or even examining hierarchies that suit user needs is often time-consuming if not intractable.

The gap between unsupervised hierarchical clustering algorithms and task-relevant user needs calls for a visual analysis solution that involves users in the hierarchy building process~\cite{liu2018bridging}. 
Our analysis of current challenges to fill the gap leads to the identification of two key requirements for improving the initial algorithmically constructed hierarchies and reducing user efforts in refining them.
First, we need to improve the quality of the initial hierarchy by combining different sources of information, for example, open-domain knowledge from publicly available ontologies and private knowledge from users.
Second, interactive refinement of the hierarchies needs to be supported and guided with visual cues to make the effort less laborious.
The visual cues are tightly coupled with the working mechanisms of the algorithms to guide the attention of users to parts of the hierarchy that the algorithms are uncertain about.

In this paper, we present an interactive steering method, ReVision, which enables users to visually supervise and steer hierarchical clustering.
Our method meets the two aforementioned requirements by augmenting unsupervised data-driven algorithms with both public and/or private knowledge.
In particular, we make the following three contributions.

First, we propose \textbf{\normalsize \xitingrevision{a constraint extraction method} based on ant colony optimization}. 
This \xitingrevision{enables us to combine} different sources of information for building an initial hierarchy of high quality.
The resulting hierarchy captures both the original data distribution (data-driven) and public and/or private knowledge from a knowledge base (knowledge-driven), such as Wikipedia.
The key challenge to achieving this contribution is to effectively identify which parts of the large public ontologies are useful and how they relate to the data items to be clustered.
To solve this problem, we combine the ant colony optimization~\cite{dorigo1997ant} with beam search~\cite{ow1988filtered}, which efficiently identifies a sub-hierarchy from the ontologies (i.e., a constraint tree) that balances tree width and depth and covers most data items with high confidence.
The constraint tree is then leveraged to cluster all data items hierarchically by using the evolutionary Bayesian rose tree algorithm~\cite{wang2013mining}.

Second, we develop \textbf{\normalsize an uncertainty-aware, tree-based interactive visualization} that enables guided refinement of hierarchies facilitated by visual cues.
The uncertainties are derived based on the model confidence of the clustering results, the constraints violation, and the structure consistency.
Leveraging the uncertainty-aware visualization, users can quickly identify the parts of the hierarchy that could benefit from steering and improve them if needed (user-driven).

Third, we present quantitative evaluation and a case study to demonstrate that ReVision facilitates the construction of a high-quality customized hierarchy based on user needs.

%% file: related.tex
\section{Related Work}
\label{sec:related-work}

\subsection{Hierarchical Clustering}
Existing work on hierarchical clustering can be divided into two categories, based on whether the constructed structures are binary (each internal node has at most two children) or multi-branch (each internal node can have more than two children). 

Pioneer \textbf{\normalsize binary} hierarchical clustering methods are metric-based.
They measure cluster similarities based on metrics such as Euclidean distance.
The metrics are used as guidance to iteratively merge similar clusters (agglomerative methods)~\cite{sibson1973slink,defays1977efficient} or split a cluster into two dissimilar sub-clusters (divisive methods)~\cite{kaufman2009finding}. 
Heller et al.~\cite{heller2005bayesian} successfully formulated hierarchical clustering as a statistical problem and obtained the best structure by maximizing the marginal likelihood function. 
Compared with traditional metric-based methods, this method has advantages in predictive capability, accuracy, and overfitting avoidance.
% and its ability to avoid overfitting.
However, the binary structure limits its application.
In \wenwen{practice}%a practical situation
, the binary structures usually fail to provide a correct and meaningful hierarchy~\cite{blundell2010bayesian}. 

\textbf{\normalsize Multi-branch} hierarchical clustering methods have been proposed to tackle the aforementioned issues. 
For example, Blundell et al. extended Bayesian hierarchical clustering~\cite{heller2005bayesian} to Bayesian rose tree (BRT)~\cite{blundell2010bayesian}, which removes the binary structure restriction. 
A greedy algorithm is applied to accelerate the clustering process.
Zavitsanos et al. developed an algorithm for text clustering based on hierarchical Dirichlet processes~\cite{zavitsanos2011non}.
Siddique and Akhtar extracted topics (long-lasting subjects) and proposed a topic-based hierarchical summarization algorithm to help users understand the topics~\cite{siddique2019topic}.
Knowles et al. proposed Pitman Yor Diffusion Tree~\cite{knowles2015pitman}, which generalizes the Dirichlet Diffusion Tree~\cite{neal2003density} to support multi-branch structure building.
Song et al.~\cite{song2015automatic} applied $k$NN-Approximation and $\varepsilon$NN-Approximation to reduce the complexity from $O(n^2\log n)$ to $O(n\log n)$, making it more applicable in practical text clustering.
While these methods are effective for building a multi-branch hierarchy that fits the data distribution, they lack a mechanism to incorporate domain knowledge. 
%Thus, 
\wenwen{As a result,}
it is difficult for the constructed hierarchies to adapt to different application scenarios.

To better incorporate domain knowledge, \textbf{\normalsize constraint-based} methods have been developed.
Constraints in the form of ``must-link'' and ``cannot-link'' have been introduced to capture scenarios that two data items must or cannot appear in the same cluster~\cite{davidson2009using,miyamoto2011constrained}.
Since the must-link and cannot-link constraints ignore
hierarchical information (the parent-child relationships), these methods may fail to reconstruct an optimized tree.
To address this issue, triple-wise constraints, which define whether two data items must be merged before the other data items merge with either of them, are presented.
An example is evolutionary Bayesian rose trees~\cite{wang2013mining}, in which the authors consider the structure extracted at the previous time as the constraints and apply them when clustering the documents at the current time point.
However, it is unclear how triple-wise constraints can be used to incorporate different types of domain knowledge (e.g., open domain knowledge and user knowledge).
In this paper, we bridge this gap by proposing a constraint extraction method based on ant colony optimization.
Moreover, we develop an uncertainty-aware interactive visualization to enable guided refinement of hierarchies.

\subsection{Visual Cluster Analysis}
Visual clustering analysis has become a research topic in the visualization community due to the abundance of clustering methods and the often noisy nature of the clustering results. 
Early research focused on the visualization of clustering results and enabling cluster comparison, while more recent research enables users to sift through large combinations of clustering parameter space and dynamically steer the clustering results~\cite{liu2014survey,liu2018bridging}. 

The Hierarchical Clustering Explorer~\cite{Seo2002HCE} is an early example that provides an overview of hierarchical clustering results applied to genomic microarray data and supports cluster comparisons of different algorithms. 
To help evaluate the quality of clusters, Cao et al. introduced an icon-based cluster visualization named DICON \cite{Cao2011DICON}, which leverages statistical information to facilitate users to interpret, evaluate, and compare clustering results. 
In a similar vein of comparison of clustering algorithms, Lex et al. introduced Matchmaker~\cite{Lex2010} to allow users to freely arrange data dimensions and compare multiple groups of them that can be clustered separately.

In addition to supporting comparison between clustering algorithms, other interactive visual clustering analysis methods enable users to steer the clustering analysis by providing feedback on data items/clusters and algorithm parameters. 
Nam et al.~\cite{Nam2007ClusterSculptor} noted that results from unsupervised clustering algorithms rarely agree with expert knowledge and intuition on the classification hierarchies. 
They therefore developed ClusterSculptor to interactively tune parameters of k-means clustering 
based on a visualization in high-dimensional space.
More recently, Cavallo et al. developed Clustrophile 2~\cite{Cavallo2019Clustrophile} to support guided exploratory clustering analysis. Given user expectations and analysis objectives, Clustrophile 2 provides guidance for users to select parameters for clustering and evaluate the quality of the corresponding results.

Many systems have been designed to improve the clustering results by interacting with data items or clusters. VISTA~\cite{VISTA} employs a star-coordinator representation to visualize multi-dimensional datasets, and it allows users to evaluate and improve the structure of the clusters with operations such as splitting and merging clusters. 
iVisClustering~\cite{lee2012iVisClustering}, an interactive document clustering system, provides both document-level and cluster-level interactions for refining the clustering results.
Clustervision~\cite{Kwon2018Clustervision} is a visual analysis system that provides quality metrics that permits users to rank and compare different clustering results. It also enables users to apply their domain knowledge to steer the analysis. Specifically, users can set up constraints to steer clustering results by establishing must-links and cannot-links for sets of data items. 
However, such constraints only work for flat clustering and do not capture the parent-child relationships in hierarchical clustering. 

Another thread of research leveraged the visualization of hierarchical topics for text data analysis, with the topic models serving as the means for clustering the text documents \cite{dou2013hierarchicaltopics,cui2014hierarchical,liu2015online}. 
Dou et al. presented HierarchicalTopics, a visual analytics system that visually present topic modeling results in a hierarchical fashion to facilitate the analysis of a large number of topics \cite{dou2013hierarchicaltopics}. The hierarchy of topics in HierachicalTopics was derived after the topics were extracted in an unsupervised fashion. 
Cui et al. proposed RoseRiver, a visual analytics system for exploring how hierarchical topics evolve in text corpora \cite{cui2014hierarchical}. 
Building on Cui et al.'s research, Liu et al. presented an online visual analysis approach to help users explore hierarchical topic evolution in text streams~\cite{liu2015online}. 
They presented a tree-cutting model to address the challenges of visualizing streaming data with a changing hierarchical topic tree layout.

Our work differs from previous research in that we leverage both public knowledge (e.g., Wikipedia) and private knowledge to improve the hierarchical clustering results.
The knowledge is captured by a set of constraints such as triples and fans and is incorporated into the hierarchical clustering process. 
In addition, an uncertainty-aware visualization is developed to help users identify where their input for improving the model is most needed.

%% file: motivation-design-system.tex
\section{Design of ReVision}\label{Sec:Design}

In this section, we first analyze the design requirements and then provide an overview of ReVision. 

\subsection{Design Requirements}
The design of ReVision was inspired by previous work on constrained clustering and hierarchical clustering.
The ReVision prototype was developed through an iterative process, during which we collaborated with two machine learning experts, including a research scientist (E1) from Microsoft who majored in interactive machine learning and a senior engineer (E2) with the Microsoft Bing News team. 
Both experts self-identified as having considerable experience in building document hierarchies.
For example, as part of E1's work, he organizes papers of interest into a folder hierarchy.
To identify recent research trends, each time a visualization or machine learning conference is held,
E1 manually refines the hierarchy to include papers from the new proceeding. 
E2 has \shixia{experience with constructing} an evolving hierarchy for news articles by using automatic clustering methods.
Each cluster in the hierarchy is considered an event.
The major events are provided to the editors, who decide the spotlight events for each day.
Both E1 and E2 consider their current practices challenging.
In particular, manually maintaining a paper hierarchy is time-consuming, while a news hierarchy built automatically is error-prone.

Based on the analysis of existing technical challenges on hierarchical clustering and discussions with the experts, the following requirements have been derived for building a hierarchy that \shixia{meets} user needs.

\textbf{R1. Automatically build a high-quality initial hierarchy by leveraging different sources of information.} % (knowledge-driven and data-driven)
Previous studies have shown that identifying an appropriate hierarchical structure is essential for exploratory data analysis~\cite{blundell2010bayesian,liu2018steering,song2015automatic}.
The experts also expressed the importance of a high-quality initial hierarchy.
E1 mentioned that automatically building the initial hierarchy was essential for achieving scalability.
\shixia{The experts further indicated that to ensure quality and suit application needs}, two types of information, knowledge and
\wenwen{intrinsic data structure, }%distribution of documents, 
need to be jointly considered.

\textit{R1.1 Integrate public and/or private knowledge when constructing the initial hierarchy (knowledge-driven).}
To improve quality for specific applications, both experts agreed that it is important to leverage knowledge embedded in existing hierarchies appropriately.
For example, \shixia{E1 indicated that it would be important to consider his existing paper hierarchy (private knowledge) and ensure the consistency of the new hierarchy with the old one. }
E2 expressed that an initial news hierarchy should fit certain existing hierarchies, for example, the taxonomy of Yahoo news or the open domain taxonomy of Wikipedia (public knowledge).

\textit{R1.2 Ensure that the constructed hierarchy summarizes the data according to the intrinsic data structure (data-driven).}
Both experts agreed that a good hierarchy should adequately capture the intrinsic data structure, which facilitates better understanding and retrieval of desired information.
For example, E2 said, ``\emph{We also want to detect new events that inherently exist in the news collection, but are not described in existing hierarchies such as Wikipedia.}''
To this end, we need to build the hierarchy according to the inherent data distribution.

\textbf{R2. Refine the hierarchy interactively based on user needs (user-driven)}. 
When comprehending a large data collection, the experts often need to effectively examine the hierarchy, identify potentially incorrect \weikai{nodes} \wenwen{(i.e., clusters of data items)}, and modify the hierarchy efficiently when needed.

\textit{R2.1 Examine and compare the \weikai{hierarchies} at multiple levels of detail.}
Both experts need to examine the hierarchies from the high-level \weikai{nodes} to the low-level descendant \weikai{nodes}.
E1 said,
``\emph{Visually examining the \weikai{hierarchies} at multiple levels is necessary to fully understand the data collection.}''
Both experts also expressed the need to compare the \weikai{hierarchical clustering results} with the constraint hierarchy for better steering the \weikai{clustering results}. % I does not change it here
For example, if the expert found that some important constraints were violated, s/he would want to compare the two hierarchies carefully and \wenwen{identify} the root cause of such violation.\looseness=-1

\textit{R2.2 Identify uncertain sub-trees.} 
The system should provide visual cues to help users quickly locate the sub-trees that need attention. 
The experts said that they were particularly interested in contradictions between existing knowledge and \wenwen{intrinsic} data distribution.
It is also desirable that the visual cues be coupled with the working mechanisms of the clustering algorithms, %for example, 
\wenwen{in order to}
reveal parts of the hierarchies that the algorithms are uncertain about.

\textit{R2.3 Modify hierarchies directly by interacting with the \weikai{nodes} and data items.}
The experts expressed the need to modify the hierarchy both at the \weikai{node}-level and item-level.
For example, E1 said that he often needed to remove some irrelevant papers and add extra papers with similar topics. 
E2 indicated that modifications of nodes (news events) were constantly needed in his work to maintain the hierarchy of the growing news collection.
Inspired by semantic interactions designed by Endert et al.~\cite{endert2012semantic}, we allow users to directly adjust \weikai{nodes} and data items (e.g., add items, merge nodes) and update the clustering results according to user feedback.

\subsection{System Overview}

The aforementioned requirements motivated us to develop a visual analysis system, ReVision, to help users build a high-quality hierarchy by combining the strengths of knowledge-driven, data-driven, and user-driven methods.
In our work, we take \textbf{textual data} as an example to illustrate the basic idea of the developed method. 
\weikai{
For example, in Fig.~\ref{fig:teaser-overview}, the constraint and clustering trees are constructed from the academic papers on ``interactive machine learning.'' 
The AMiner Science Knowledge Graph~\cite{data_aminer}, a graph that organizes Computer Science academic papers based on the ACM computing classification system, is utilized as the knowledge base. 
More details about the data and knowledge base can be found in Sec.~\ref{sec:application}}.
ReVision can also be applied to other types of data as long as the similarity between two data items can be measured.

\begin{figure}[!htb]
\centering
\includegraphics[width=\linewidth]{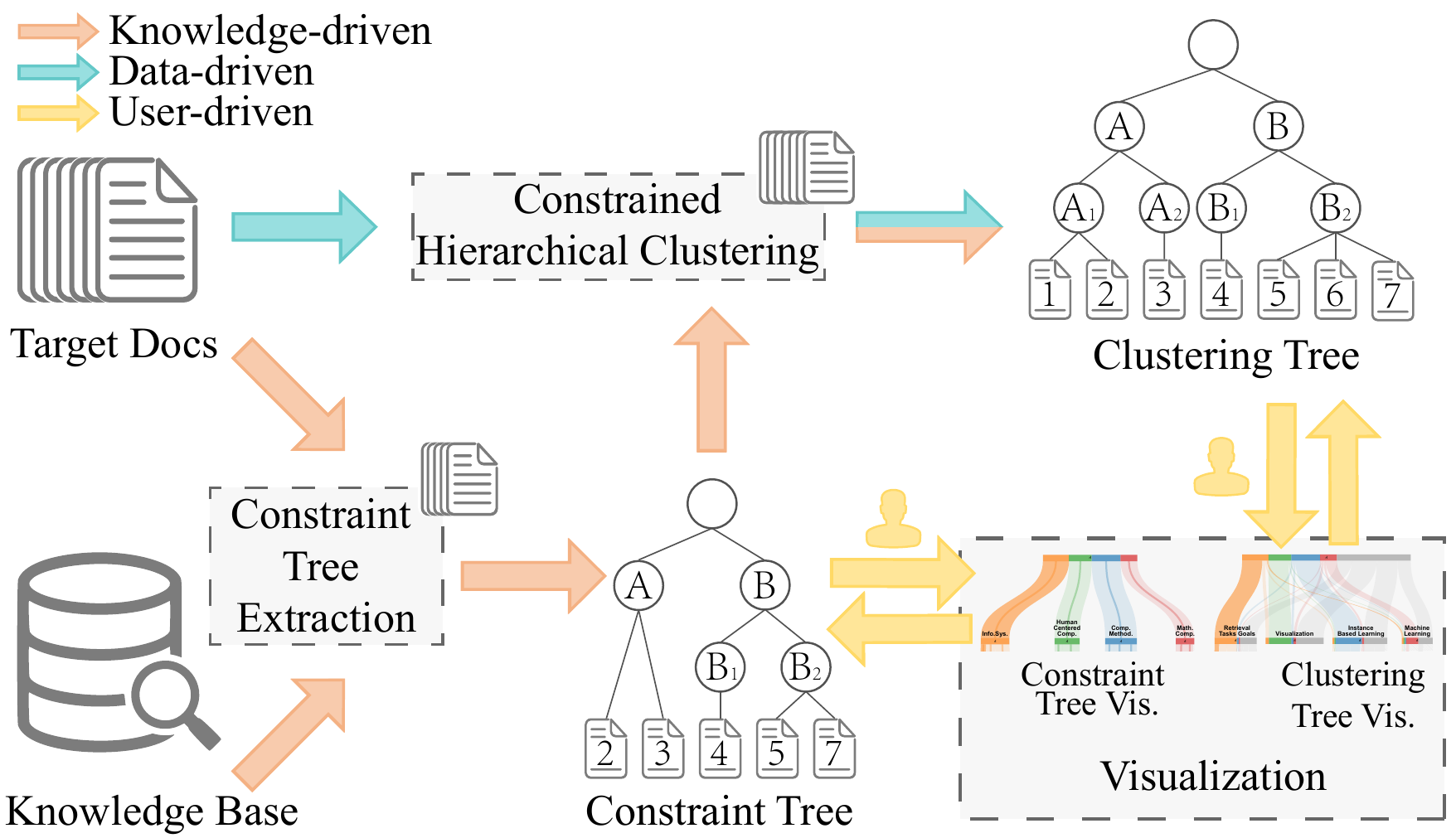}
\caption{
The pipeline of ReVision. Given the documents to be clustered, we build the constraints using the knowledge base (knowledge-driven).
\weikai{These constraints %only consider 
\wenwen{are constructed with} a subset of the documents.}
The constraints \wenwen{are then} applied to guide the clustering process \weikai{for all documents} (knowledge- and data-driven).
Users can modify the constraint tree and the clustering tree to meet their customized needs through the visualization (user-driven).
}
\label{fig:pipeline}
\end{figure}

The ReVision system consists of two major parts: 1) \wenwen{a} hierarchical clustering \wenwen{method} that builds a high-quality tree based on knowledge and data distribution (\textbf{R1}); and 2) a tree-based visualization that helps refine the hierarchy based on user needs (\textbf{R2}) (Fig.~\ref{fig:pipeline}).
Hierarchical clustering consists of two components. 
The first component, constraint tree extraction, identifies which parts of the knowledge are useful for guiding the construction of the \textbf{constraint tree} (\textbf{R1.1}).
\shixia{This constraint tree only consists of a subset of the documents that are mapped to part of the knowledge base with high confidence. }
The second component, constrained clustering, builds the hierarchy by considering both the extracted constraints (\textbf{R1.1}) and data distribution (\textbf{R1.2}).
\weikai{It takes all the documents and constraints as the input and generates the \textbf{clustering tree}.}
\xitingrevision{
The tree-based visualization facilitates the comparison of hierarchies with the juxtaposition approach and consistent color encoding (R2.1). 
This visualization is designed to facilitate the identification of uncertain sub-hierarchies (R2.2) and the modification of the hierarchy at the node level (R2.3).
The information panel helps with the examination of documents and keywords of a node (R2.1) and enables the modification at the document level (R2.3).}

\begin{figure}[!htb]
\centering
\includegraphics[width=\linewidth]{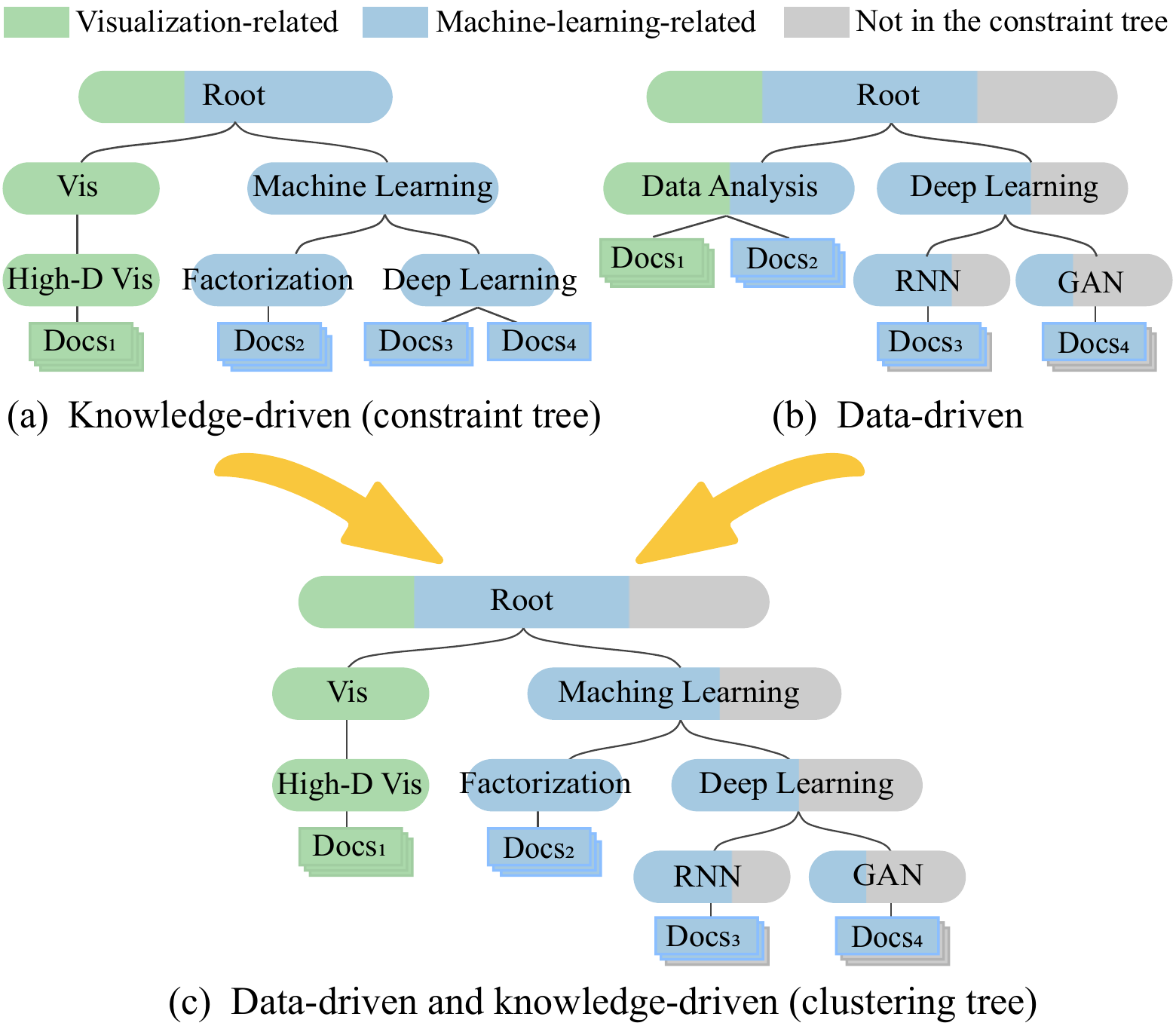}
\caption{\weikai{
Illustrate the relationships among a knowledge-driven \wenwen{only} method, a data-driven \wenwen{only} method, and the proposed method.
The rectangles represent the documents linked to the leaf nodes, and the rounded rectangles represent the nodes in the hierarchy.
The color of the documents is assigned based on the first-level categories of the constraint tree (a), and the colors of the nodes show the proportions of the documents \wenwen{of different categories. }%with different colors.
}}
\label{fig:toy}
\end{figure}

%% file: data-modeling.tex
\section{Hierarchical Clustering}
In this section, we first introduce the overall procedure for hierarchical clustering. 
Then we illustrate the constraint tree extraction method.

\subsection{Algorithm Overview}

\shixia{Before introducing the algorithm in detail, we use a simple example (Fig.~\ref{fig:toy}) to illustrate the benefits of combining data-driven and knowledge-driven methods. 
In the constraint tree (Fig.~\ref{fig:toy}(a)), documents about \xitingrevision{``High-D Vis'' (High-Dimensional Visualization)} are successfully separated from ``Factorization,'' 
which are mixed in the data-driven result under the node ``Data Analysis'' (Fig.~\ref{fig:toy}(b)) since they share many keywords, such as ``multivariate'' and ``clustering.'' 
The data-driven clustering result can extract the intrinsic structure from among the documents about ``RNN'' and ``GAN.'' 
However, these two nodes are not covered by the knowledge base as it only contains a node ``Deep Learning.''
\xitingrevision{Moreover, the constraint tree only contains a subset of documents.
The other documents cannot be included due to their low similarity with existing nodes in the knowledge base.}
By combining the advantages of the knowledge-driven and data-driven methods, the clustering tree (Fig.~\ref{fig:toy}(c)) provides a better result.}\looseness=-1

Motivated by the above example, the hierarchical clustering method builds a high-quality initial hierarchy by considering two types of information: 1) public and/or private knowledge represented by an existing hierarchy (a knowledge base) (\textbf{R1.1}) and 2) distribution of the data items to be clustered (\textbf{R1.2}).
Accordingly, our method contains two steps: \textbf{constraint tree extraction} and \textbf{constrained hierarchical clustering}.

\shixia{
According to Wang et al.~\cite{wang2013mining}, the constrained hierarchical clustering problem can be solved by maximizing the posterior probability of \xitingrevision{the constructed hierarchy} $T$
\begin{equation}
p(T\mid D, T_c) \propto p(D\mid T)p(T\mid T_c).
\label{eq:posterior}
\end{equation}
Here, $p(D\mid T)$ represents how well $T$ fits the distribution of \xitingrevision{the data item in corpus} $D$ (data-driven) and $p(T\mid T_c)$ denotes how similar $T$ is to \xitingrevision{the constraint tree} $T_c$ (knowledge-driven).
}
\xitingrevision{By considering each data item as an initial sub-tree, Eq.~(\ref{eq:posterior}) can be optimized by using a greedy agglomerative
strategy~\cite{blundell2010bayesian}, which iteratively merges the two sub-trees that result in the highest posterior probability gain~\cite{wang2013mining}.}\looseness=-1

\shixia{In ReVision, we directly utilize this constrained clustering algorithm to hierarchically 
\xitingrevision{cluster}
textual data based on the extracted constraint tree. 
\xitingrevision{
While Wang et al. focus on evolutionary clustering and consider $T_c$ as the tree constructed at the last time point, we need to extract constraint trees based on the knowledge bases. % for the data corpus 
Our problem is more challenging due to the large scale of the knowledge bases and their document diversity.
}
As a result, we focus on introducing the constraint tree extraction method in the following subsection. 
}
\subsection{%Basic Idea of 
Constraint Tree Extraction}
\label{Alg.Ant}

Constraint tree extraction aims to identify which parts of a large knowledge base are useful and how they relate to the data items to be clustered ({\textbf{R1.1}}).
Given a knowledge base, which is usually a directed acyclic graph (DAG), the constraint tree is a sub-hierarchy of the knowledge base that is relevant to the documents to be clustered.
The constraint tree should have related documents assigned to the most suitable \weikai{nodes} (accuracy and coverage) and have a succinct and balanced structure (structure simplicity).

A straightforward method to construct the constraint tree is to map each document to the most similar node in the knowledge base.
However, there could be multiple good candidates in the knowledge base for a document.
If we consider each document independently, we could easily obtain a constraint tree with many \weikai{nodes} scattered across the knowledge base.
To determine which candidate is the best, we need to take into account the distribution of relevant documents.
Take a document about ``Soccer Rules'' as an example. It is more similar to the \weikai{node} ``Sports Rules'' than to ``Soccer.''
However, if many of the documents to be clustered are about ``Soccer'' rather than ``Sports Rules,'' assigning it to ``Soccer'' will be a better choice.

Based on this observation, we propose an ant-colony-based method that extracts the constraint tree by considering the overall document distribution.
In particular, 
the \textbf{ant colony optimization}~\cite{dorigo1997ant} is adopted to obtain a more accurate and succinct hierarchical structure.
\textbf{Beam search pruning}~\cite{ow1988filtered} is also applied to quickly locate the most promising parts of the knowledge base and speedup the constraint tree extraction.

\subsubsection{Ant Colony Optimization}
The ant colony optimization algorithm is a probabilistic technique for finding good paths in a graph. 
We leverage this algorithm to find a constraint tree that consists of good paths by considering documents as ants.
The optimization algorithm consists of two steps: 1) projection of documents; and 2) extraction of constraints.

\textbf{Projection of documents}. 
In this step, each document to be clustered is mapped to multiple candidate nodes in the knowledge base.
Given a set of documents $D=\{d_1,\ldots,d_n\}$, for each $d_i$, we first retrieve $K$ documents from the knowledge base that are most similar to $d_i$.
\xiting{This is achieved by using an open-source search engine, Apache Lucene~\cite{bialecki2012apache}.}
\weikai{It decomposes document $d_i$ into a set of query words and then retrieves the documents that contain these words by using an inverted index.
The inverted index is an index data structure that stores a mapping from words to nodes in the knowledge base. 
It makes the retrieval of relevant documents more efficient.
Then we calculate the similarity scores for all $n \times K$ \wenwen{document} pairs and keep the top $q$ percent of them.
}
%\shixia{If document $d_i$ occurs in a kept document pair $\langle d_i, k_i\rangle$, it will be projected to $k_i$. 
%Here $k_i$ is a document in the knowledge base.}
\weikaiminor{If document $d_i$ occurs in a kept document pair $\langle d_i, d'_j\rangle$, it will be projected to $d'_j$. 
Here $d'_j$ is a document in the knowledge base.}
%\weikai{The similarity score is calculated based on the cosine similarity of the vector representations of two documents, which are extracted using the word embedding model~\cite{mikolov2013wordembedding}.}
\weikaiminor{The similarity score is calculated based on the cosine similarity of the vector representations of two documents.
The vector representation of each document is computed by averaging the word vectors, each of which is extracted by using a pre-trained word embedding model~\cite{mikolov2013wordembedding}.}
\weikai{Empirically, we set $K=50$ since it is sufficient for various constraints to be extracted later.}
The sensitivity of $q$ is discussed in Sec.~\ref{sec:exp_ant}. 
\shixia{Based on the sensitivity analysis, we set $q=10\%$ in our implementation.}

\begin{figure}[b]
\centering
\includegraphics[width=\linewidth]{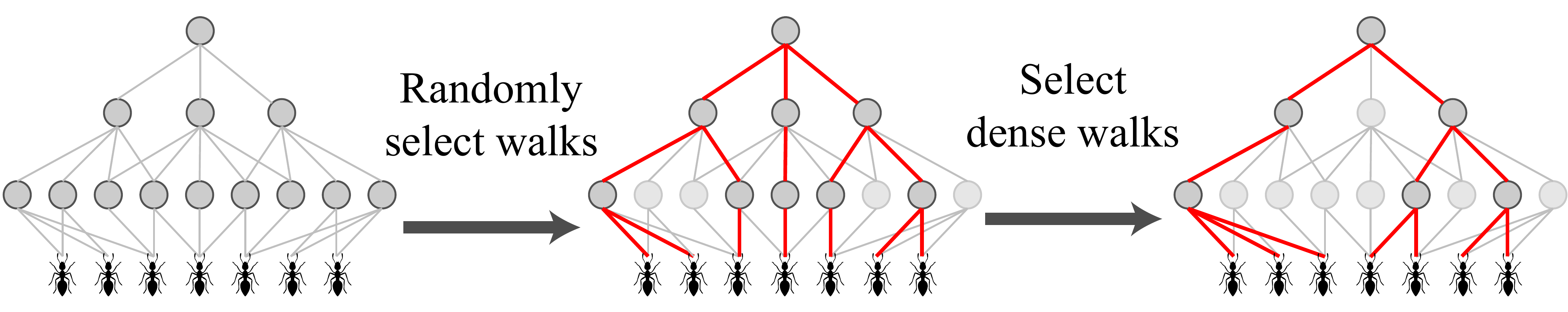}
\put(-225,-8){\small(a)}
\put(-131,-8){\small(b)}
\put(-36,-8){\small(c)}
\caption{An example shows how the ant colony optimization extracts the constraints: (a) the ants randomly select the walks with the same probability;
(b) the ants select the walks based on the pheromone.
Here, the third ant joins the first two ants because the walk of the two ants has a larger pheromone; the forth one joins the fifth one because it has a larger pheromone, too.
}
\vspace{-3mm}
\label{fig:Ant colony}
\end{figure}

\textbf{Extraction of constraints}. 
We then extract the constraint tree based on the idea of the ant colony algorithm.
In particular, each projected document is considered to be an ant.
In each iteration, the ants move gradually from the projected documents to the upper-level nodes based on the information (pheromone) other ants leave.
The ants have a larger probability of going to nodes with a higher pheromone.
All ants cooperate to find an appropriate constraint tree, which consists of the ants' walks to the root.
\weikai{A walk contains a set of nodes as well as the edges connecting them}.
Fig.~\ref{fig:Ant colony} explains how the algorithm works by using a simple example.

The key problem here is to determine the pheromone $\tau_{\overline{uv}}$.
Here $u$ is a node in the knowledge base, $v$ is one parent of $u$, and $\overline{uv}$ is the edge connecting $u$ and $v$.
The probability that an ant at $u$ goes to $v$ is proportional to $\tau_{\overline{uv}}$:
$p_{\overline{uv}}=\tau_{\overline{uv}}/\sum_{v'\in\mathrm{parent}(u)}\tau_{\overline{uv'}}$.
Note that many knowledge bases (e.g., Wikipedia) are DAGs, so it is natural for $u$ to have multiple parents.
We first initialize $\tau_{\overline{uv}}$ to be evenly distributed: $\tau_{\overline{uv_i}}=\tau_{\overline{uv_j}}, \forall v_i, v_j\in \text{parent}(u)$.
In each iteration, $\tau_{\overline{uv}}$ is updated to reveal information found by ants that pass this node.
In the original ant colony algorithm, $\tau_{\overline{uv}}$ is updated according to walk length $L$ \weikai{(the number of the edges in the walk)}.
The ants lay down more pheromones on shorter walks so that they will aggregate on the shortest walks after several iterations.
In our scenario, finding the shortest walks is not enough. We also need to simultaneously consider accuracy $A$, coverage $R$, and structure simplicity $S$.
\xitingrevision{To consider these factors simultaneously, we follow the original ant colony algorithm, which increases the pheromone of the paths with desirable properties at the beginning of each iteration}
\begin{equation}
\tau_{\overline{uv}}^{(i+1)}=\rho\tau_{\overline{uv}}^{(i)}+ARS\label{eq:new_tauuv_}.
\end{equation}
Here, $0<\rho<1$ denotes how fast the pheromone evaporates and is usually set to $0.9$. 
In our implementation, we define the accuracy, coverage, and structure simplicity measures as follows.

\textbf{Accuracy ($A$)} ensures that the path fits the projected documents.
The fitness can be measured by Dirichlet compound multinomial (DCM) distribution~\cite{liu2012automatic},
which calculates the probability that a node generates a document according to the word distribution of the document and node.
An intuitive idea is to define $A$ as the average generative probability:
$A = \sum_{v\in w} \log f_{\text{DCM}}(d,v)/L$.
Here $d$ denotes the projected document of the ant, $v$ is a node on walk $w$, $L$ is the length of walk $w$, and $f_{\text{DCM}}(d,v)$ represents the probability that node $v$ generates document $d$ according to the DCM distribution~\cite{madsen2005modeling}.
A larger $f_{\text{DCM}}(d,v)$ indicates that $d$ better fits $v$.
However, according to this definition, we have $A<0$, which may affect the convergence of $\tau_{\overline{uv}}$.
To tackle this issue, we use the transformation $f(x)=-1/x$ to convert the negative value to a positive one while preserving its monotonicity
\begin{equation}
A=-\frac{L}{\sum\limits_{v\in w}\log f_{\text{DCM}}(d, v)}.
\label{eq:accuracy}
\end{equation}

\textbf{Coverage ($R$)} encourages choosing a node where most of its children are covered by the documents.
Ensuring coverage can avoid selecting large but meaningless \weikai{nodes}, such as ``Category:Living people'' in Wikipedia, which helps little in information understanding.
The coverage is defined as 
\begin{equation}
R = \min_{v\in w} \frac{n_v'}{n_v}.
\label{eq:coverage}
\end{equation} 
Here, $n_v$ denotes the number of child nodes of node $v$, and $n_v'$ denotes the child number of $v$ that is visited by at least one ant.

\textbf{Structure simplicity ($S$)}.
A succinct and balanced structure should be neither too deep nor too wide.
\xitingrevision{The depth of the tree can be punished by defining $S$ as $1/L^\gamma$.
Here, $\gamma>0$ controls the depth of the extracted constraint hierarchy.
The larger the $\gamma$ is, the shallower the tree depth is. 
As a result, a larger $\gamma$ is usually preferred for knowledge bases with deeper structures, e.g., Wikipedia.
In such cases, we usually need to obtain a shallower structure by using a larger $\gamma$. 
}
\xitingrevision{We further modify $S$ to consider the overall size of the structure}.
To this end, we define a density metric $\frac{1}{L}\sum_{\overline{uv}\in w}N(\overline{uv})\label{eq:density}$.
Here, $\overline{uv}$ is an edge in walk $w$, and $N(\overline{uv})$ represents the number of ants that pass edge $\overline{uv}$.
The density metric ensures that we only include a node in the constraint tree if there are sufficient ants that pass this node.
The final structure simplicity is defined by combining the \xitingrevision{tree depth} penalty and density
\begin{equation}
S = \frac{1}{L^{\gamma+1}}\sum_{\overline{uv}\in w}N(\overline{uv})\label{eq:simplicity}.
\end{equation}

\subsubsection{Beam Search Pruning}
The ant colony optimization can be very slow when applied to a large knowledge base such as Wikipedia.
The most computationally expensive step is accuracy calculation (Eq.~(\ref{eq:accuracy})), in which we load the term-frequency vectors of nodes and calculate $f_{\text{DCM}}(d,v)$.
To improve computational efficiency, we propose a top-down beam search method to quickly identify the region of interest and prune irrelevant nodes.

Our pruning method is developed based on the observation that only a very small part of the large knowledge base is relevant to the documents to be clustered.
For the irrelevant parts, the accuracy values are usually quite small and can be set to a small constant value.
Thus, we only update the accuracy values for relevant \weikai{nodes}, which are found by using a top-down beam search.
Specifically, we first find the most relevant \weikai{nodes} at the first level by using the following voting function
\begin{equation}
 \text{vote}(v) = \sum_{d\in D}\frac{f_{\text{DCM}}(d, v)}{\sum_{v'\in V}f_{\text{DCM}}(d,v')},
 \label{eq:beamsearch}
\end{equation}
where $V$ is the candidate node set.
The children of the selected \weikai{nodes} are considered to be candidates in the next level.
We enumerate the candidates to find the $k$ most relevant \weikai{nodes} at the next level by using Eq.~(\ref{eq:beamsearch}) again.
The above process is performed iteratively so that for every level of the knowledge base, we only calculate accuracy for the $k$ most relevant \weikai{nodes}.

%% file: visualization.tex
\section{Revision Visualization}\label{sec:vis}
To meet the design requirements discussed in Sec.~\ref{Sec:Design}, we develop an uncertainty-aware tree visualization, which allows users to explore the hierarchical constraints and \weikai{clustering results} (\textbf{R2.1}), examine the overall constraints satisfaction (\textbf{R2.2}), and modify the hierarchy based on users' requirements (\textbf{R2.3}).

\subsection{\weikai{Hierarchy} as Node-Link Diagram}
\label{subsec:VisualDesign}

\subsubsection{Visual Design}
To visually illustrate the \weikai{hierarchy}, we choose the node-link diagram since it is intuitive and shows the hierarchical structure clearly (\textbf{R2.1}).
As shown in Fig.~\ref{fig:teaser-overview}(b)(c),
the constraint tree and the \weikai{clustering tree} are placed in juxtaposition to facilitate the comparison.
In the constraint tree (Fig.~\ref{fig:teaser-overview}(b)), a node represents a \weikai{set of documents}, a link encodes a parent-child relationship (i.e., \weikai{sub-nodes}), and the color assignment is determined by its first-level \weikai{nodes}.
\xitingrevision{
To facilitate the comparison between the \weikai{clustering tree} and constraint tree, we use identical colors to denote the same sets of documents.
For example, in Fig.~\ref{fig:teaser-overview}(c), ``Retrieval Tasks Goals" has an orange branch because some of its documents belong to ``Information Systems," which is colored orange in the constraint tree (Fig.~\ref{fig:teaser-overview}(b)). 
The \weikai{clustering tree} is visualized similarly to the constraint tree except \weikai{for} two major differences.
First, the gray color is used to encode the document groups without constraints.
Second, a parent-child relationship may be split into multiple colored stripes according to how the constraints distribute on it.}
The width of a node encodes the number of documents in it.
The label of each node $v$ is set as the name of the corresponding node in the knowledge base, which has the highest generative probability to $v$.
Different types of nodes are marked with different glyphs: $\vcenter{\hbox{\includegraphics[height=1.5\fontcharht\font`\B]{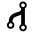} }}$\! denotes internal nodes, $\vcenter{\hbox{\includegraphics[height=1.5\fontcharht\font`\B]{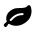}}}$ represents leaf nodes, and \includegraphics[height=1\fontcharht\font`\B]{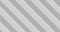} represents collapsed nodes.
\xitingrevision{Some nodes are collapsed in the initial visualization due to the limited screen space.
The collapsed nodes are automatically computed based on a tree cut algorithm~\cite{furnas1986generalized}.}
As an example shown in Fig.~\ref{fig:teaser-overview}(C), a subset of children under the “Machine Learning” branch in the clustering tree are collapsed.

\subsubsection{Layout}
\label{subsec:VisualLayout}
The layout algorithm for the node-link diagram consists of two steps: 
1) \textit{node ordering} that balances readability, the similarity between adjacent nodes, and stability between consecutive layouts;
and 2) \textit{tree cutting} that provides informative \weikai{nodes} for user examination.

\noindent\textbf{Node ordering}.
Node ordering is very important to generate a legible and informative tree layout.
In our method, three factors, similarity, readability, and stability, are considered for producing a layout with good ordering.

\textit{Similarity}.
The similarity factor aims to place nodes with similar content close to each other in a tree to facilitate exploration.
We employ the cost function of a state-of-the-art ordering algorithm,
optimal leaf ordering~\cite{bar2001fast}, which maximizes the similarity of adjacent node pairs

\begin{equation}
\text{similarity}(\sigma)=-\sum_{i=2}^{n}\text{similarity}(v_{\sigma_{i-1}},v_{\sigma_i})\label{eq:similarity}.
\end{equation}
\weikai{Here, $\sigma$ is an ordering of $(1,\ldots,n)$, $\sigma_i$ is the index of the node at position $i$, and
$v_{\sigma_i}$ is the $i$th node in the given ordering.
}
The similarity between two nodes is calculated \wenwen{as} the cosine similarity of the node vectors, which are extracted using word embedding~\cite{mikolov2013wordembedding}.\looseness =-1

\textit{Readability}.
To improve readability, we aim to reduce visual clutter by minimizing edge crossings. 
The edge crossings occur when the constraint distribution is displayed on the \weikai{clustering tree} (Fig.~\ref{fig:teaser-overview}(c)).
In our implementation, we only consider the constraint distribution of the first-level categories in the constraint tree, whose order is already determined.
Assume we have $m$ categories and $n$ children for ordering.
Let $n_{ij}$ be the number of documents in child $v_i$ with category $j$. If $v_i$ is placed before $v_j$, the crossing cost between them is
\begin{equation}
\mathrm{cross}(v_i,v_j)=\sum_{k=1}^{m}\sum_{l=k+1}^{m} n_{il}n_{jk}\label{eq:readability-single},
\end{equation}
and hence the total crossing cost is defined as 
\begin{equation}
\text{readability}(\sigma)=\sum_{i=1}^{n}\sum_{j=i+1}^{n}\mathrm{cross}(v_{\sigma_i},v_{\sigma_j})\label{eq:readability-multi}.
\end{equation}

\textit{Stability}.
The readability cost changes if users adjust the structure of the tree, causing a change of ordering as well.
However, users might get confused if the order of nodes changes dramatically after certain adjustments.
As a result, we add one more term in the cost function to maintain stability.
We formulate stability as preserving the relative distance to the previous ordering

\begin{equation}
\text{stability}(\sigma)=\sum_{i=1}^{n}\sum_{j=i+1}^{n}\lvert (i-j) - (\text{pos}(v_{\sigma_i})-\text{pos}(v_{\sigma_j}))\rvert,
\label{eq:stability}
\end{equation}
where $\text{pos}(v_{\sigma_i})$ is the position of $v_{\sigma_i}$ in the previous order.

With all the cost function outlined, the challenge is to optimize all of them together.
Not only is the search space exponential, but the costs also have contradictory properties.
For example, distance matters in similarity but not in readability; while the relative ordering is important in readability, it has no impact on similarity.
To tackle these challenges, we use simulated annealing~\cite{aarts1988simulated} to search for the optimal ordering in the discrete search space. 
By applying simulated annealing, an optimal ordering of the nodes is generated for the node-link diagram visualization.

\noindent\textbf{Tree Cutting}.
Displaying all the nodes is difficult due to the limited space, and hardly ever necessary due to limited attention. 
Thus, it is necessary to dynamically show a part of the tree that users are interested in. 
We adopt a degree-of-interest (DOI)~\cite{furnas1986generalized} strategy to determine which nodes to display.
The nodes with high DOI value are more likely to be displayed on the screen.
DOI is calculated as 
\begin{equation}
\mathrm{DOI}(v) = \mathrm{API}(v) - D(v,f),
\end{equation}
where $\mathrm{API}(v)$ is the prior interest of node $v$, 
$f$ is the focus node chosen by the user, 
and $D(v,f)$ is the tree-distance between node $v$ and $f$.
\shixia{Since users tend to be more interested in large nodes with high uncertainty, the prior interest of each node is set as the product of the document number of that node and its uncertainty value.
}
In case the tree cutting algorithm predicts certain nodes unimportant but the user would still like to keep them visible, we provide a pin-down function \includegraphics[height=\fontcharht\font`\B]{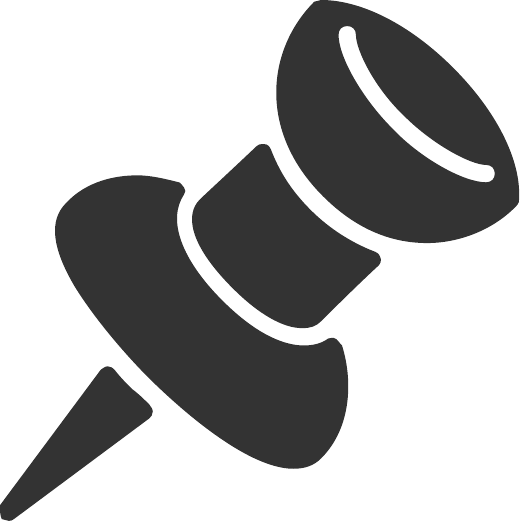}. 
This function allows the user to explore more nodes while keeping the nodes of interest in context.

\subsection{Uncertainty as Line-Based Glyph}
\label{subsec:uncertainty}
\subsubsection{Visual Design}
\begin{wrapfigure}[7]{r}{0.232\textwidth}
 \vspace{-12pt}
\includegraphics[width=0.232\textwidth]{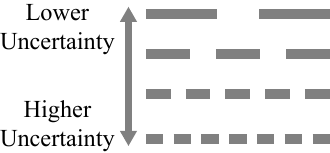}
\caption{\label{fig:uncertainty}The grain glyph.}
\end{wrapfigure}
To better guide user efforts in refining the hierarchical structure, it is essential to provide visual cues that highlight parts of the hierarchy that exhibit a higher degree of parent-child relationship uncertainty, which is measured by the combination of model-related, knowledge-related, and structure-consistency uncertainty (\textbf{R2.2}).
To encode the uncertainty of the link connecting the node and its parent in the constraint tree, we used the grain glyph (Fig.~\ref{fig:uncertainty}), which has shown to be effective at encoding uncertainty in graph edges~\cite{guo2015representing, bertin1983semiology}.
As shown in Fig.~\ref{fig:uncertainty}, the grain glyph uses the variation of fineness or coarseness of dashes to encode the degree of uncertainty. 
Specifically, links with coarser dashes indicate a higher degree of uncertainty.
In Fig.~\ref{fig:teaser-overview}(A), the link connecting node ``WWW'' (World Wide Web) and its parent node ``Information Systems'' in the constraint tree appears in coarser dashes, indicating a higher degree of uncertainty.

\subsubsection{Modeling}
To help users better locate uncertain sub-structures, we compute an overall uncertainty score for each node. The score is calculated as the weighted average of the following three uncertainty factors.

\xitingrevision{\textit{Model-related uncertainty} is measured by using $-\log p(T\mid D, T_c)$, where $p(T\mid D, T_c)$ is the posterior probability defined in Eq.~(\ref{eq:posterior}).
A higher model-related uncertainty score reflects lower confidence in the constraint structure.
}

\textit{Knowledge-related uncertainty}
is measured by the contradiction between the public domain knowledge and the inherent distribution of the documents.
The contradiction is caused by the dispersion phenomenon where the documents belonging to one node in the constraints may be distributed into many nodes in the \weikai{clustering tree} and vice versa.
As a result, 
we evaluate the uncertainty using the information entropy~\cite{chowdhury2010introduction}, 
which takes into account the proportions of documents distributed into different nodes.

\textit{Structure-consistency uncertainty} measures content compatibility of parent-child relationships in the clustering hierarchy.
We use the hierarchies in Fig.~\ref{fig:teaser-overview} as an example to illustrate the concept of the content compatibility.
It consists of many visual-analytics-related papers. 
In the initial constraint tree, ``Information System'' contains three child nodes, ``Information Retrieval," ``Information Systems Applications," and ``World Wide Web'' (Fig.~\ref{fig:teaser-overview}(A)).
Most of the papers in ``Information Retrieval" are related to the retrieval of information from document repositories.
``Information Systems Applications" is an academic study of systems and applications for collecting, filtering, processing, creating, and distributing data. 
Both nodes are compatible with their parent node, namely, ``Information System."
Most of the papers in ``World Wide Web" focus on the research of social media analysis.
However, some papers about graph layout are also included in this \weikai{node}, which are not relevant to ``Information System.''
Thus, ``World Wide Web" is less compatible with ``Information System."
The compatibility problem can be formulated as a set-inclusion between a node and its child node.
Here a node is regarded as a set, and a document is considered an element of the set.
In the fuzzy set theory~\cite{smithson2006fuzzy}, subsethood is the degree of containment of one set in another.
Accordingly, we leverage the concept of subsethood to calculate the structure-consistency uncertainty.
Given a child node $v_c$ and its parent $v_p$, we define the subsethood for ${v_c}\subset{v_p}$ as 
\begin{equation}
\alpha_{v_c,v_p} = \dfrac{\sum_{d}\min\{\mu_{{v_c}}(d),\mu_{{v_p}}(d)\}}{\sum_{d}\mu_{{v_c}}(d)},
\label{eq:subsethood}
\end{equation}
where $\mu_{{v_c}}(d)$ is a membership function calculated by cosine similarity \weikai{of the vector representations of} a document $d$ and a node ${v_c}$, indicating how well $d$ belongs to ${v_c}$. 
\xitingrevision{
$\alpha_{v_c,v_p}$ will be $1$ if $\mu_{{v_p}}(x)\ge \mu_{{v_c}}(d)$ holds for all $d$.
The structure-consistency uncertainty is set as $(1-\alpha_{v_c,v_p})\in [0,1]$.} %R3Q18

The overall uncertainty of a node is a weighted average of model-, knowledge-, and structure-consistency uncertainty.
In our implementation, the weights are set to 1, 3, and 4, respectively.

\subsection{User Interactions}
\label{subsec:VisualInteractions}
To facilitate interactive refinement of the hierarchy, ReVision supports the following interactions (\textbf{R2.1, R2.3}).

\noindent\textbf{Hierarchy-level exploration (\textbf{R2.1}).}
To facilitate the exploration of the hierarchy, we allow users to expand all children of a node by double-clicking the node.
The constraint tree and the \weikai{clustering tree} are coordinated. 
For example, when a user clicks a node in the constraint tree, the nodes in the \weikai{clustering tree} that contain the same documents as in the constraint tree node are highlighted. 
Such coordination facilitates users in tracking and comparing \weikai{nodes} between the two hierarchies.
In addition, users can get detailed information (label, word cloud, and document list) by clicking a node.\looseness=-1

\noindent\textbf{\weikai{Node}-level refinement (\textbf{R2.3}).}
Multiple user interactions, including merging, removing, and hierarchy rebuilding are developed to support \weikai{node}-level refinement. 
To merge two nodes, a user can drag a node and choose one of three options to merge it with another node.
The three merging options are absorb\, $\vcenter{\hbox{\includegraphics[height=1.6\fontcharht\font`\B]{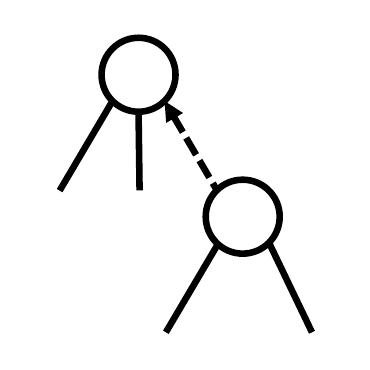}}}$\! (one node is absorbed as a child of the other), join\, $\vcenter{\hbox{\includegraphics[height=1.6\fontcharht\font`\B]{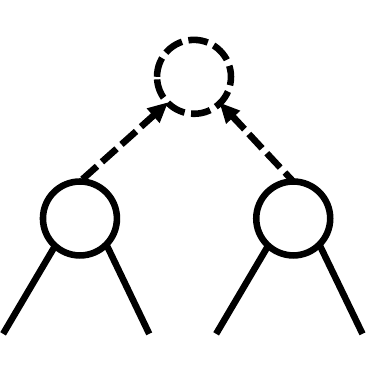}}}$ (a newly-created node with two nodes as children), and collapse\!
$\vcenter{\hbox{\includegraphics[height=1.6\fontcharht\font`\B]{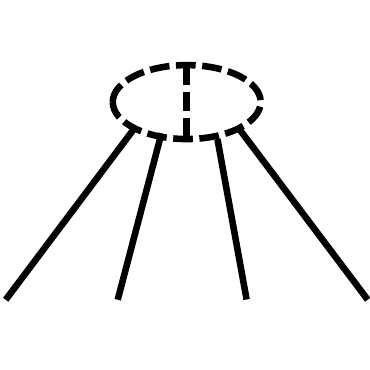}}}$ (children of the two nodes are merged together).
The user can also cancel\!
$\vcenter{\hbox{\includegraphics[height=1.5\fontcharht\font`\B]{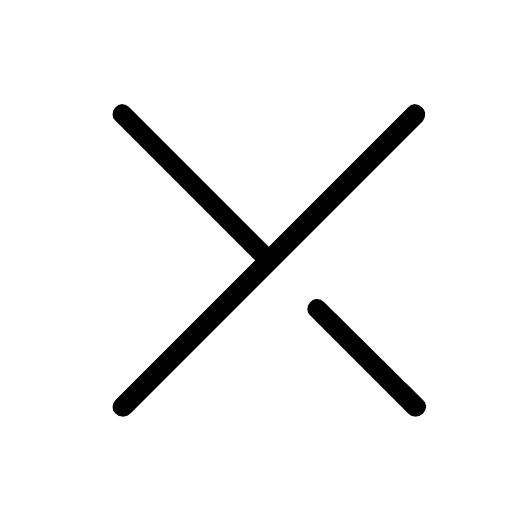}}}$\! the merging operation.
For nodes that are considered irrelevant, the user can remove them by clicking $\vcenter{\hbox{\includegraphics[height=1.5\fontcharht\font`\B]{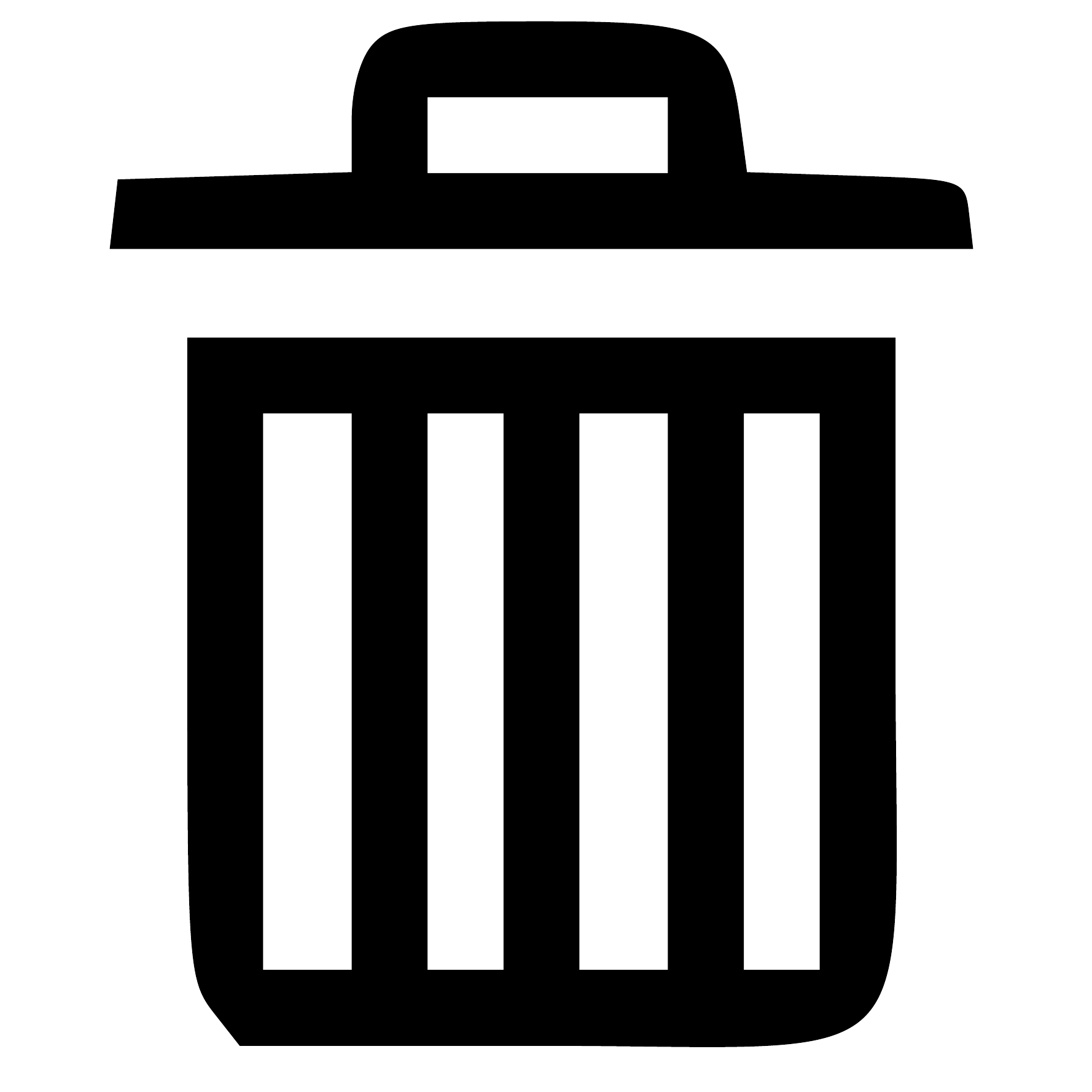}}}$ in the information panel.
If the user is not satisfied with the hierarchical structure in a sub-tree, s/he can click $\vcenter{\hbox{\includegraphics[height=1.5\fontcharht\font`\B]{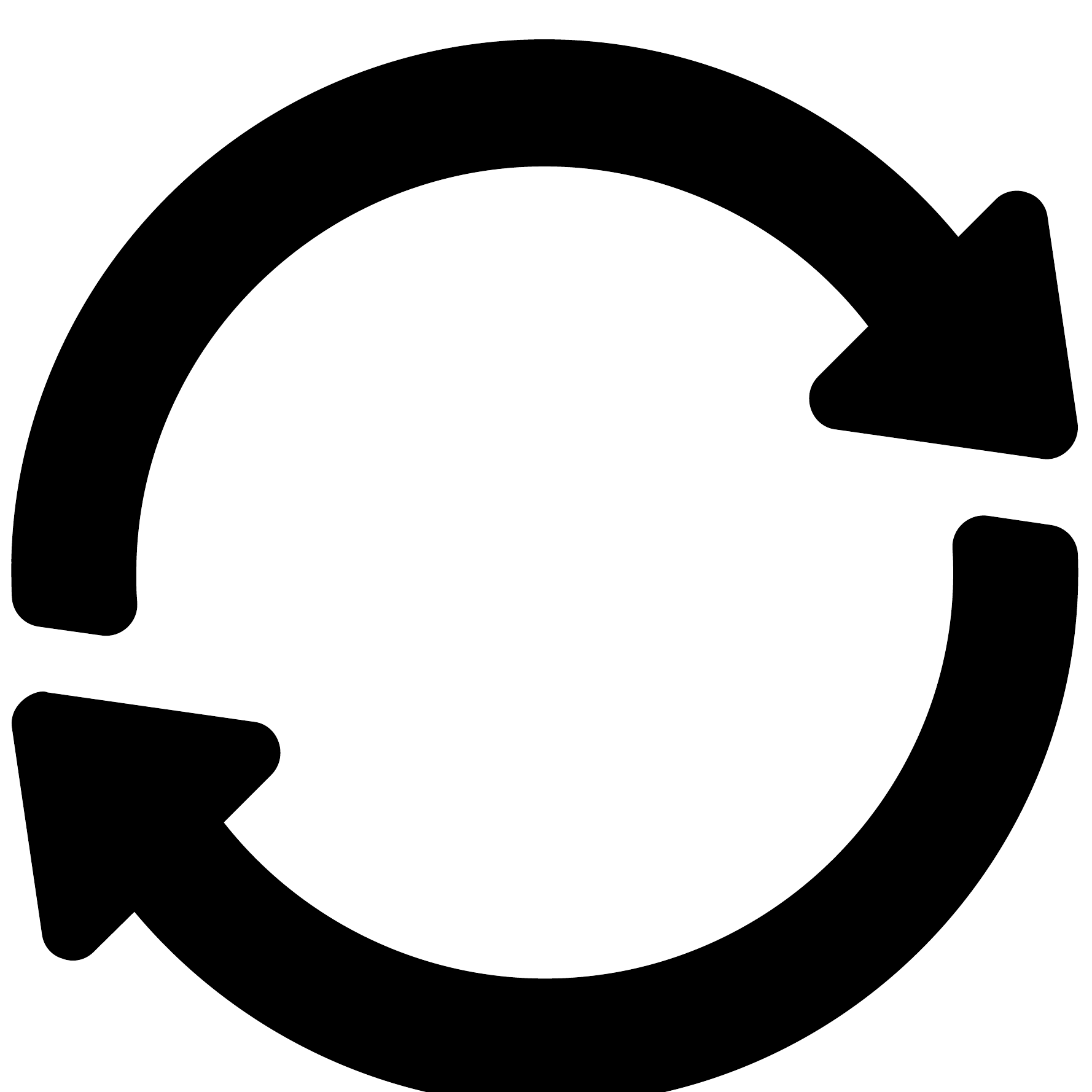}}}$ and rebuild the sub-tree by applying BRT only to the documents in the node directly.

\noindent\textbf{Document-level refinement (R2.3).}
 A user can also perform document-level modifications, such as removing irrelevant documents or moving certain documents to a more relevant \weikai{node}, to refine the hierarchy more precisely.
Specifically, a user can first click a \weikai{node} to examine the document list and identify misclassified documents.
To support the examination of a large number of documents, search ($\vcenter{\hbox{\includegraphics[height=1.5\fontcharht\font`\B]{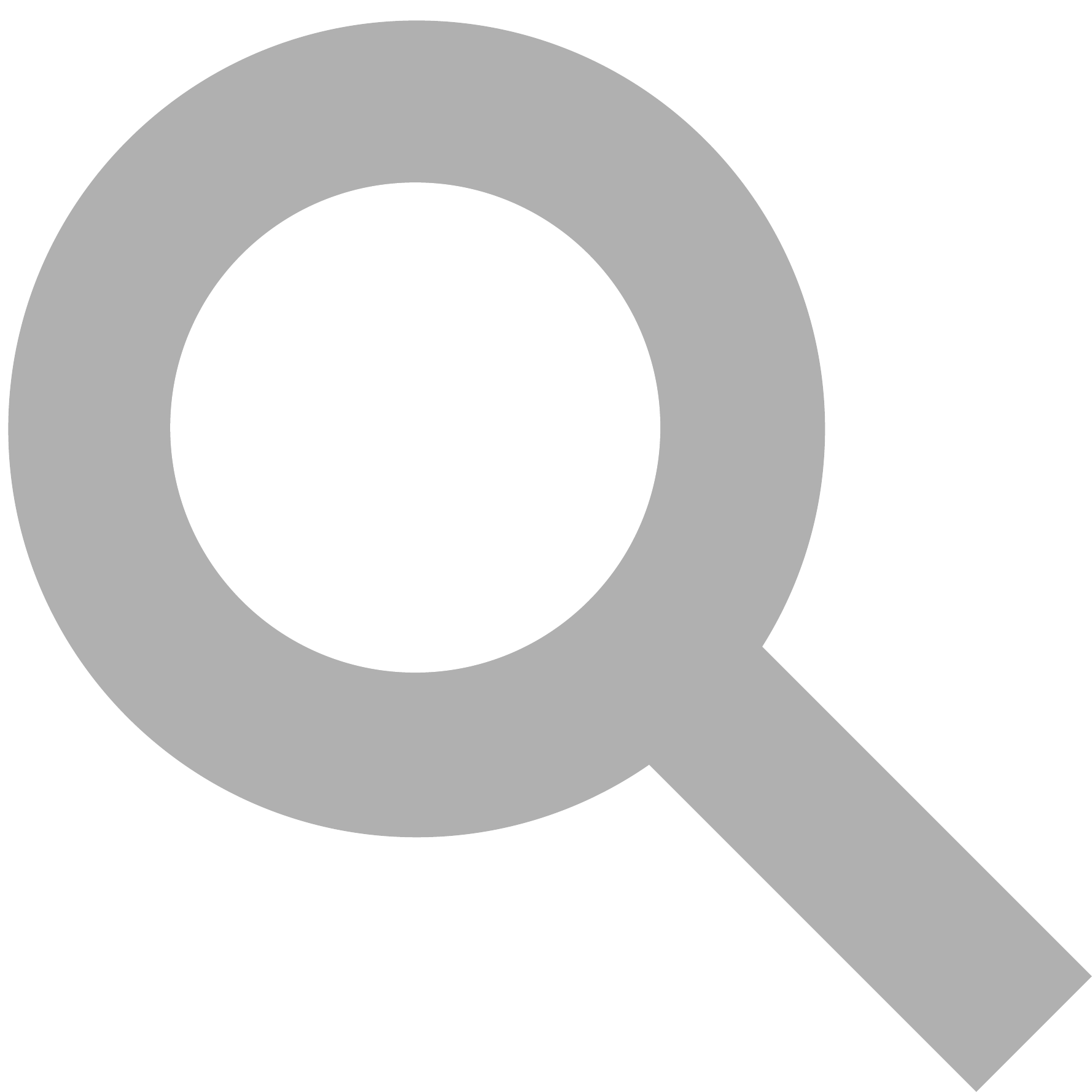}}}$) and filtering ($\vcenter{\hbox{\includegraphics[height=1.2\fontcharht\font`\B]{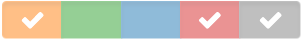}}}$) operations are provided to facilitate the identification of misclassified documents.
These documents will then be removed $\vcenter{\hbox{\includegraphics[height=1.5\fontcharht\font`\B]{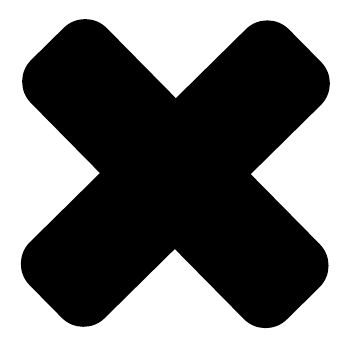}}}$ from the \weikai{node}.
If needed, the user can add other relevant documents into the \weikai{node} through the button $\vcenter{\hbox{\includegraphics[height=1.5\fontcharht\font`\B]{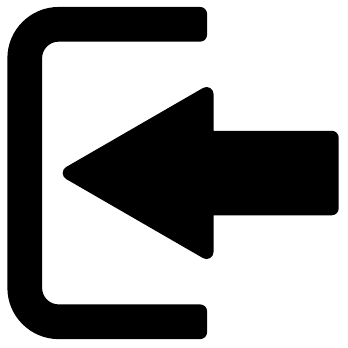}}}$ in the document list.
To facilitate the refinement of certain topics, a starred document list is provided so that users can collect the documents first, and then add them to the proper nodes later.

\wenwen{\noindent\textbf{Other user interactions and information displayed in ReVision.} 
The control panel (Fig.~\ref{fig:teaser-overview}(a)) supports loading constraints and updating a \weikai{clustering tree}. 
A user can load and add papers using the buttons in the top left corner of the interface. 
The ``Constraint Weight'' slider allows users to interactively adjust the constraint weight, 
\xitingrevision{which balances the importance of the knowledge-driven term $p(T|T_c)$ and the data-driven term $p(D|T)$ in constrained clustering (Eq.~(\ref{eq:posterior})).
A larger constraint weight usually results in a clustering tree that is more similar to the constraint tree~\cite{wang2013mining}.}
The information panel (Fig.~\ref{fig:teaser-overview}(d)) is developed to facilitate the understanding and customization of the \weikai{hierarchies}. 
For example, \weikai{the word cloud provides a summary of the high-frequency words in the documents of the selected node.
The bottom table displays the document titles.
Users can click a document title to see the detailed content.}
If s/he thinks this document is not relevant, s/he can remove this document or move it to another node. 
}
\looseness=-1

%% file: application.tex
\section{Evaluation}
\label{sec:application}

\begin{figure*}[t]
\centering
\subfigure[Category set A, $q=30\%$]{\includegraphics[width=0.245\linewidth]{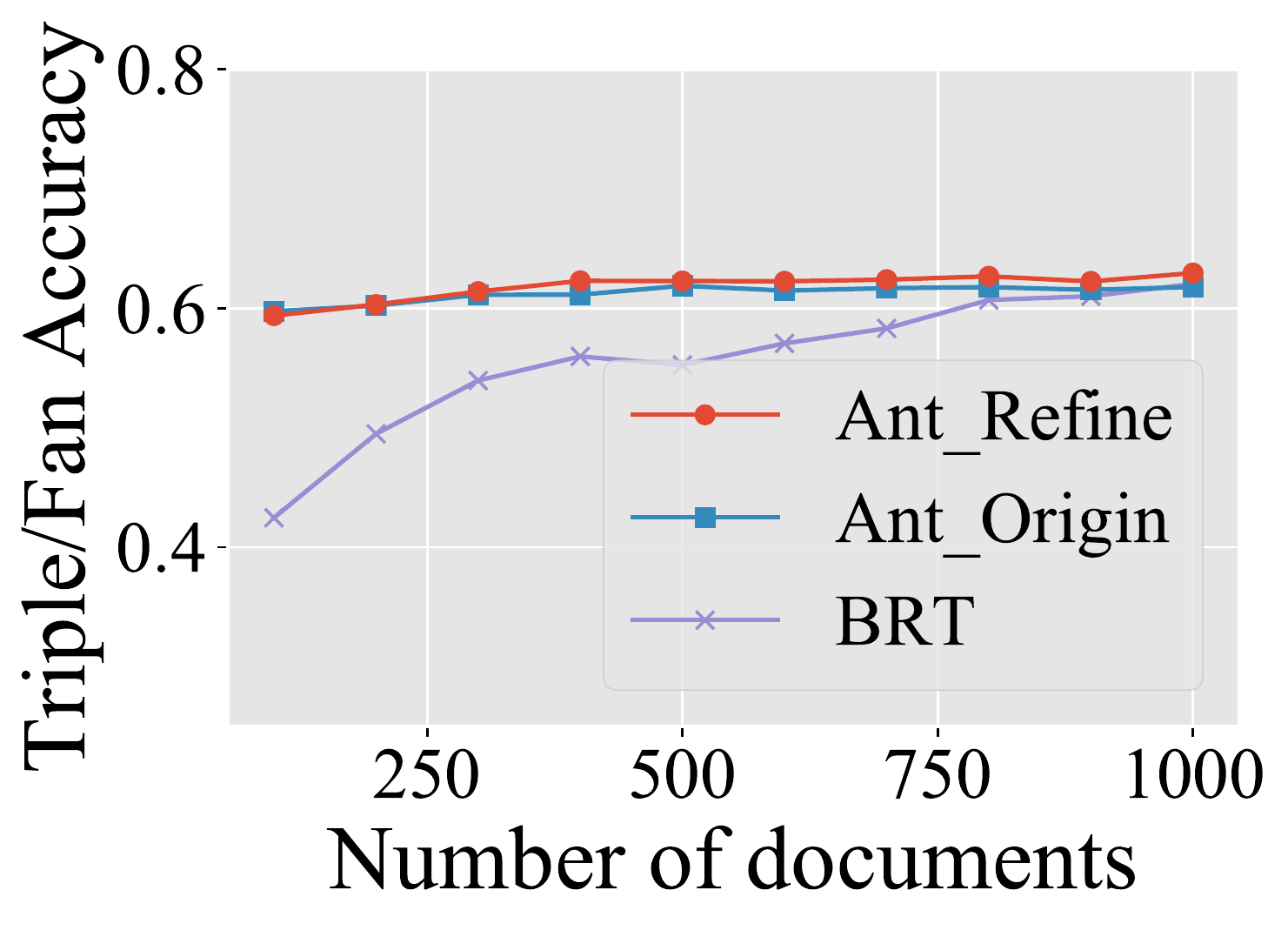}}
\subfigure[Category set A, $q=20\%$]{\includegraphics[width=0.245\linewidth]{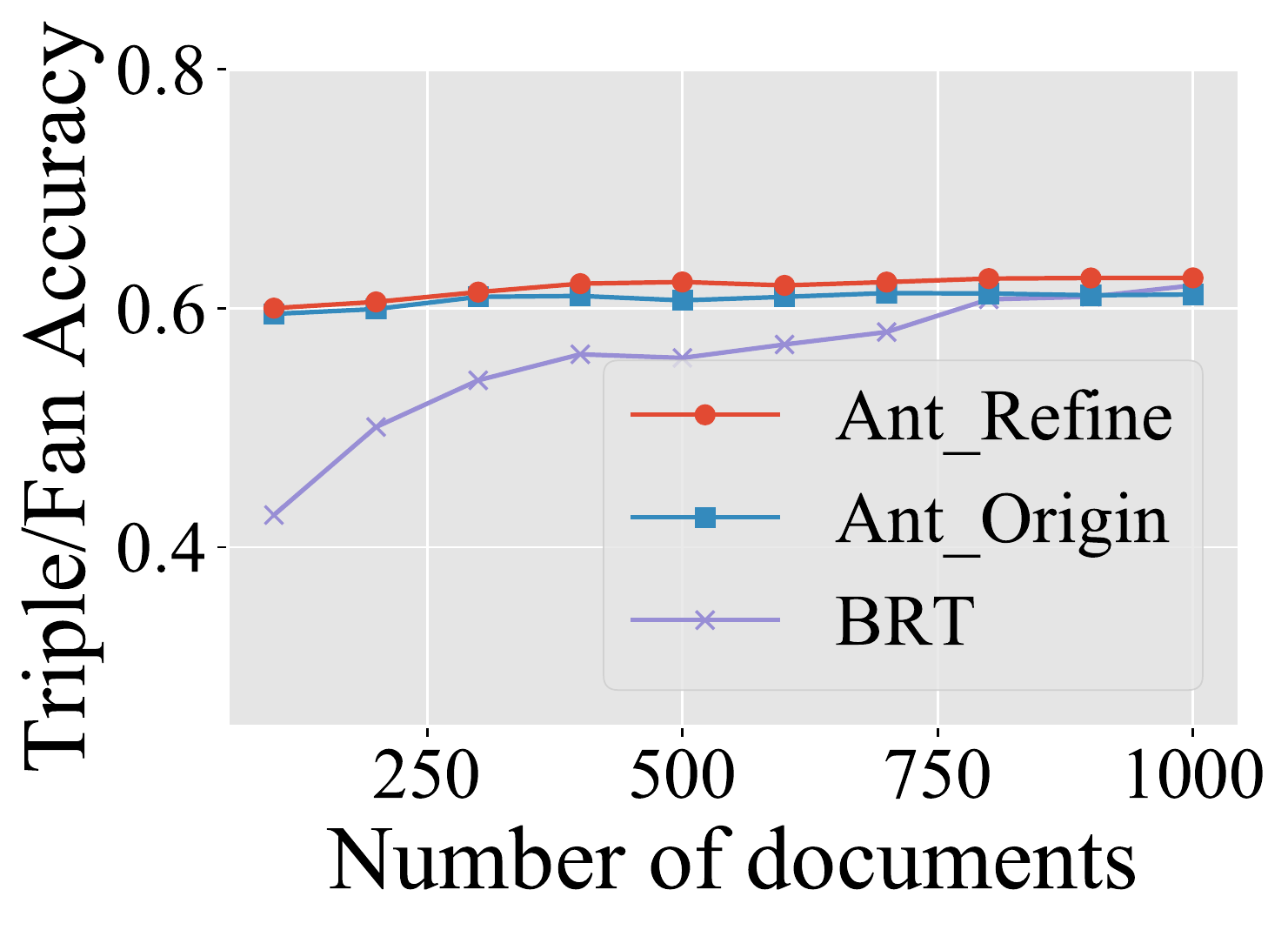}}
\subfigure[Category set A, $q=10\%$]{\includegraphics[width=0.245\linewidth]{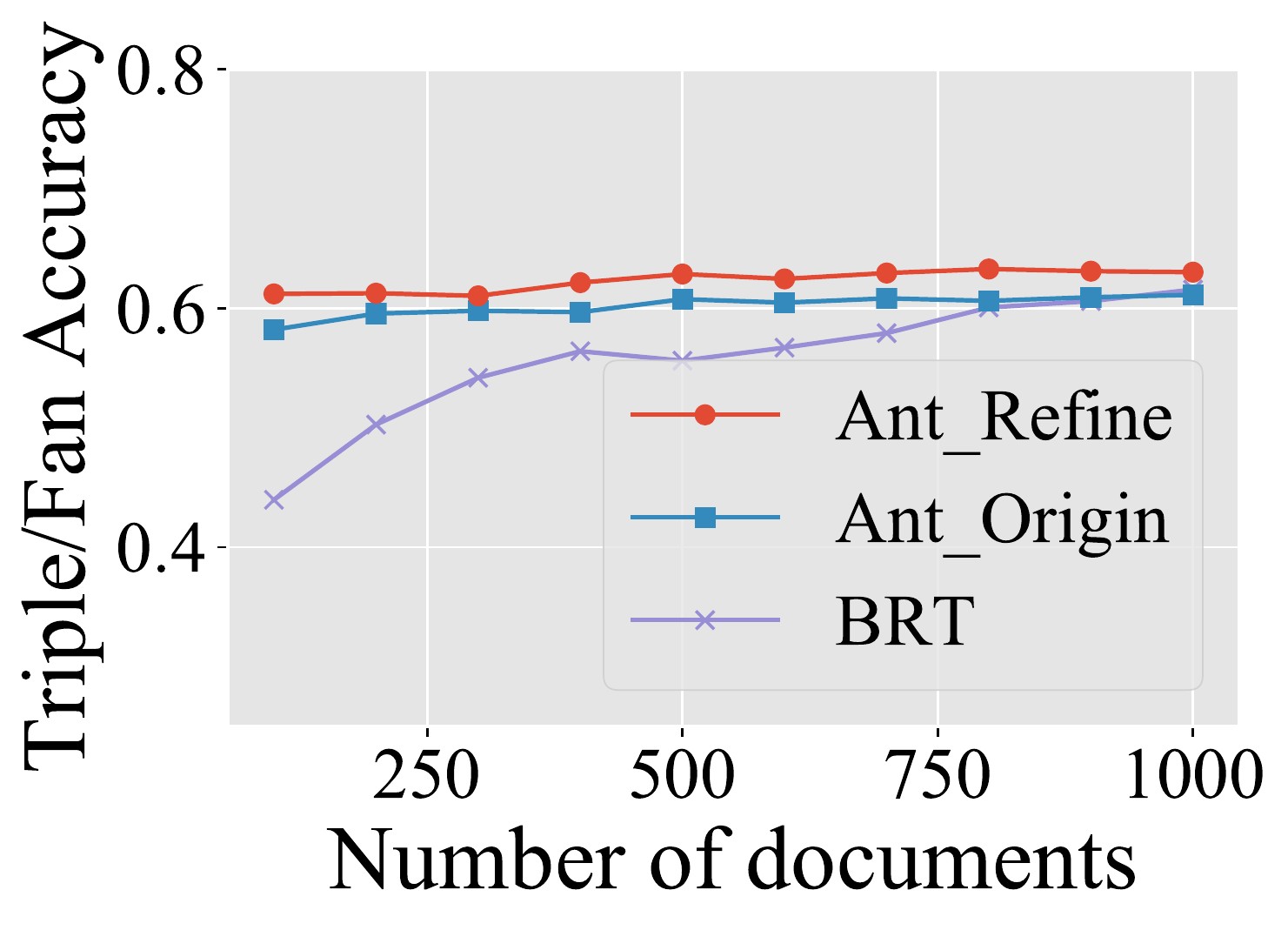}}
\subfigure[Category set B, $q=10\%$]{\includegraphics[width=0.245\linewidth]{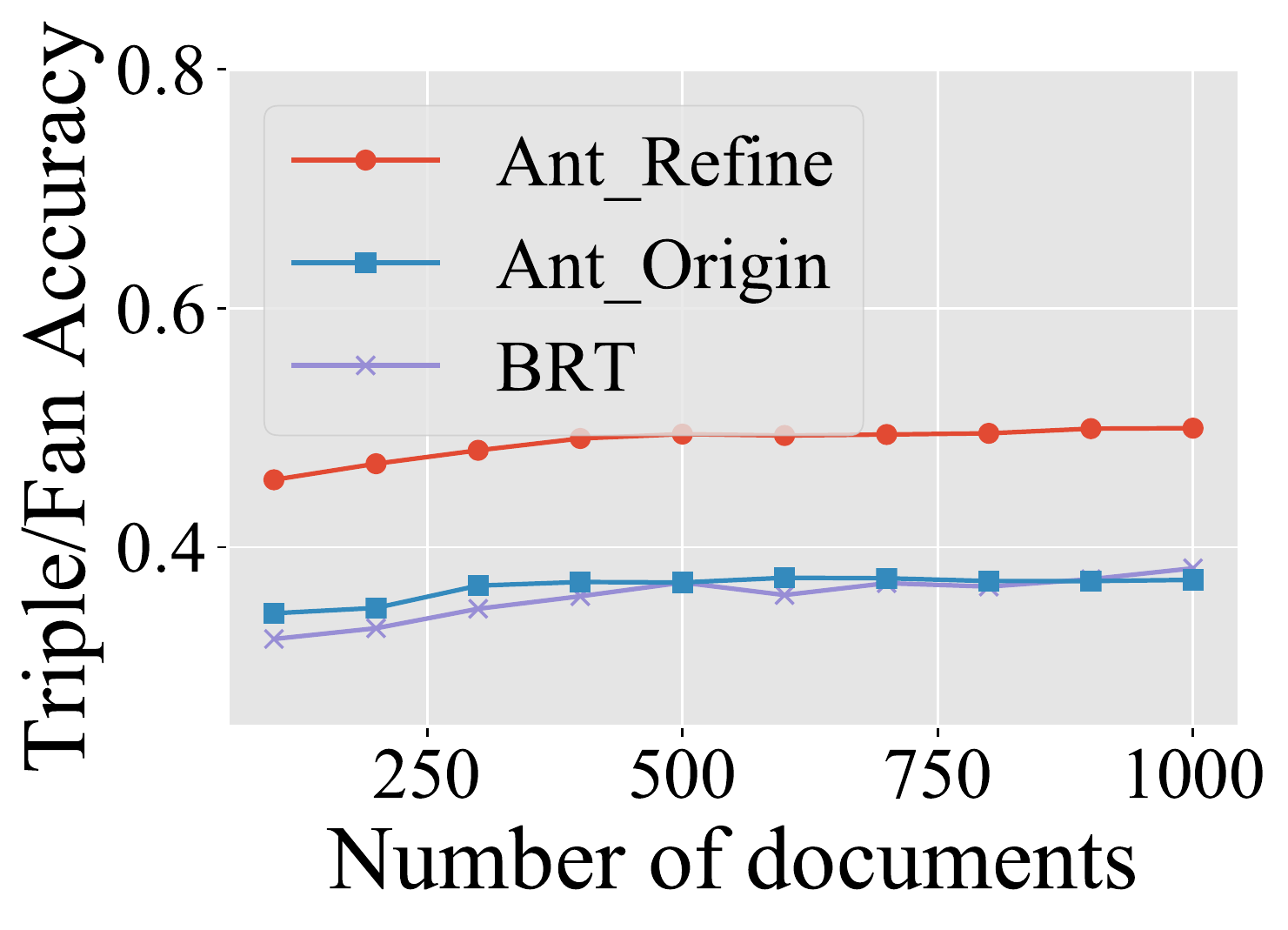}}
\caption{Comparison of constraint tree quality with different parameter settings.}
\label{fig:EXP_Part1_combine}
\end{figure*}

We conducted three quantitative experiments and a case study to evaluate the effectiveness and usefulness of our method.
The quantitative analysis demonstrates the effectiveness of the constrained hierarchical clustering algorithm.
The case study illustrates the usefulness of ReVision in helping users construct customized clustering trees.
Four datasets are used in the evaluation. 

\noindent\textbf{20 Newsgroups}~\cite{data_20news} is a collection of approximately 20,000 news articles,
\xitingrevision{
which are organized into a hierarchy with 20 categories.
The first-level contains 7 categories, and the second-level contains 16 categories.
We cleaned this hierarchy by first removing the categories that are labeled as miscellaneous, e.g., \emph{\textbf{misc}.forsale} and \emph{talk.politics.\textbf{misc}}.
The periods in the category names denote parent-child relationships, e.g., \emph{misc.forsale} indicates that \emph{forsale} is a child of \emph{misc}.
We further cleaned the hierarchy by removing categories with only one child (e.g., \emph{soc.religion}) and category pairs that are semantically similar but are placed far away from each other in the hierarchy (e.g., \emph{comp.sys.mac.hardware} and \emph{sci.electronics}). 
This results in a relatively balanced two-level hierarchy with 4 first-level categories and 9 second-level categories.
The first-level categories are \emph{comp} (computer), \emph{sci} (science), \emph{rec} (recreation), and \emph{talk}.
The second-level categories include \emph{comp.graphics}, \emph{sci.med} (medicine), \emph{sci.space}, \emph{rec.baseball}, \emph{rec.hockey}, \emph{rec.autos} (automobiles), \emph{rec.motorcycles}, \emph{talk.guns}, and \emph{talk.mideast}.
We denoted these selected categories as \textbf{category set B}.
This set contains articles that are difficult to categorize due to the existence of categories with similar content.
For example, the articles in \emph{sci.space} may also be placed in \emph{rec.autos} since they are related to transportation.
To evaluate whether our methods perform better than the baselines in constructing less complex hierarchies, we created \textbf{category set A} by removing more ambiguous categories (e.g., \emph{sci.space}).
This results in a hierarchy with four categories, i.e., \emph{comp.graphics}, \emph{sci.med}, \emph{rec.baseball}, and \emph{rec.hockey}.
}
\weikai{Datasets A and B contain 3,917 and 8,714 documents, respectively.}%R3Q21

\noindent\textbf{New York Times Annotated Corpus}~\cite{data_nyt} contains over 1.8 million articles published by the New York Times.
We chose ten categories that have an adequate number of documents and sub-hierarchies. These ten categories are split into three branches at the first level, and the numbers of the sub-categories in each branch are five, three, and two, respectively.
Approximately 83,000 documents remained in this dataset after filtering.

\noindent\textbf{AMiner Science Knowledge Graph} (SciKG)~\cite{data_aminer} is a rich knowledge graph designed for scientific purposes, 
which contains 20,000 leaf nodes.
Each leaf node contains 50 relevant papers. 
We selected 12 categories (each category contains dozens to hundreds of leaf nodes) and built a three-level hierarchy with three internal nodes in the first level. 
One internal node contains three categories directly, while the other two nodes contain two internal nodes in the second level, respectively. 
These four second-level internal nodes contain 2, 2, 2, and 3 categories, respectively.
AMiner is regarded as the public knowledge for the constraint tree extraction for academic papers.

\noindent\textbf{DBpedia}~\cite{auer2007dbpedia} is a knowledge base extracted from Wikipedia. It contains about 885,000 categories and 4,642,000 articles.
DBpedia is considered as the public knowledge for constraint tree extraction for general text such as news articles.

\subsection{Quantitative Experiments}
Three quantitative experiments were conducted to evaluate the constraints extraction and constrained clustering method.

\subsubsection{Effectiveness of Constraints Extraction}\label{sec:exp_ant}
In this experiment, we demonstrate that the constraint tree extracted by using the ant-colony-based method is accurate compared with the ground-truth hierarchy.

\noindent \textbf{Experimental settings.}
We built the constraint trees for articles in 20 Newsgroups based on DBpedia.
Three methods were evaluated.
\textbf{Ant\_Origin} is the ant-colony-based constraint tree extraction method described in Sec.~\ref{Alg.Ant}.
\shixia{\textbf{Ant\_Refine} modifies the constraint tree built by Ant\_Origin based on user refinements. 
For example, node ``Rocket'' was moved from its previous parent ``Transport'' to the new parent ``Astronomy'' (absorb), 
\weikai{and node ``Health'' was merged with node ``Medicine'' (collapse).}
The refinements usually take five to ten steps.}
\textbf{BRT}~\cite{heller2005bayesian} is the baseline method.
It builds a multi-branch hierarchy by optimizing the likelihood (the data-driven term $p(D|T)$).
Public knowledge from DBpedia is not considered in BRT.
We evaluated how the three methods perform with different hyperparameter settings and different numbers of documents.
Specifically, we tested different configurations of hyperparameter $q$ (\{10\%, 20\%, 30\%\}).
A larger \weikai{percentage} $q$ indicates that a larger number of news articles were kept in the constraint tree.
To evaluate how the methods perform on different numbers of news articles, we sampled $n$ documents ($n\in \{100,200,...,1000\}$).
To eliminate biases caused by sampling, we sampled 10 times for each $n$ and averaged the results.

\noindent \textbf{Evaluation criterion.}
We measured the quality of constraint trees by comparing them with the ground-truth hierarchy provided in the 20 Newsgroup dataset.
In particular, we evaluated the \textbf{triple/fan accuracy}, which is defined as $n_c/n_a$.
$n_c$ denotes the number of triples or fans that appear in both the constraint tree and the ground-truth hierarchy.
$n_a$ is the number of triples or fans in the ground-truth hierarchy.

\noindent \textbf{Results.}
Fig.~\ref{fig:EXP_Part1_combine} compares our method with the baseline in terms of triple/fan accuracy.
By analyzing the result, we have the following three conclusions.

\emph{Overall performance}.
Our constraint tree extraction method, Ant\_Origin, consistently performs better than BRT in both \textit{dataset A} and \textit{dataset B}, with an average improvement of 4\%.
This demonstrates that our method effectively leverages public knowledge from DBpedia to improve constraint tree quality.
Ant\_Refine consistently performs better than Ant\_Origin, which shows the usefulness of incorporating user refinements.

\emph{Effect of projection quantile $q$}.
Figs.~\ref{fig:EXP_Part1_combine}(a)--(c) show the performance of the three methods with different values of $q$. %on category set A
Our method achieved a stable result (always around 0.6) no matter which value of $q$ was used.
This demonstrates that the quality of the constraint tree built by the ant-colony-based method is not sensitive to the setting of $q$.

\emph{Effect of the document number}.
While BRT can only achieve good results when the number of documents is large, our method can achieve good results even when a few documents are provided.
Compared with BRT, Ant\_Origin improves accuracy by 12\% if only 100 news articles were given.
This is because BRT is purely data-driven.
It considers only the distribution of the news articles when building the constraint tree.
In contrast to BRT, our method also considers the public knowledge from DBpedia.
This ensures a good result even when the number of documents is small.

\subsubsection{Sensitivity Analysis in Ant Colony Optimization}
\weikaiminor{In this experiment, we conduct the sensitivity analysis of the parameters (accuracy ($A$), coverage ($R$), and structure simplicity ($S$)) used in ant colony optimization (Eq.~\ref{eq:new_tauuv_}) when extracting the constraints.
For the first two parameters, there are no more adjustable parameters in them so that the ablation study is enough to verify the sensitivity.
However, there is a parameter $\gamma$ controlling the punishment of the height of a tree in structure simplicity ($S$), so we conduct more experiment under different $\gamma$ to reveal its influence.}

\noindent\textbf{Experimental settings.}
\weikaiminor{
We used 20 Newsgroup dataset and AMiner dataset so that both the DBpedia and AMiner SciKG are used in our experiments.
The triple/fan accuracy is used to measure the quality of the extracted constraints.
For the ablation study on accuracy ($A$) and coverage ($R$), the different $\gamma$ will affect the accuracy, so we reported the best accuracy under different $\gamma\in\{0,1,2,3,4,5\}$.
For the sensitivity analysis on $\gamma$, we compared the accuracy under different $\gamma\in\{0,1,2,3,4,5\}$ to see how it affects the constraints extraction.
All the results reported are averaged on 10 trails.
}

\noindent\textbf{Results.}
\weikaiminor{
As shown in Table~\ref{table:ablation}, removing either accuracy ($A$) or coverage ($R$) will cause a decrease in the quality of the extracted constraints.
The sensitivity analysis (Table~\ref{table:sensitivity}) shows that for a deep and complex knowledge base like DBpedia, $\gamma$ plays an important role in the constraints extraction, and a larger $\gamma=4$ is needed.
However, for the knowledge base with a shallow and simple structure, the extracted constraints are not so sensitive to $\gamma$, and a smaller $\gamma=1$ can yield good results.
}
\begin{table}[htb]
%\caption{The ablation study on accuracy ($A$) and coverage ($R$).\looseness=-1}
\caption{The triple/fan accuracy in the ablation study on accuracy ($A$) and coverage ($R$) with different knowledge bases.\looseness=-1}
\label{table:ablation}
\setlength{\tabcolsep}{1.35mm}{
\begin{tabular}{ccccc}
\hline
Experiment setting & Remove both & Only A & Only R & A and R \\
\hline
DBpedia & 0.417 & 0.534 & 0.568 & 0.619  \\
AMiner SciKG & 0.451 & 0.604 & 0.613 & 0.639 \\
\hline
\end{tabular}
}
\end{table}

\begin{table}[htb]
%\caption{The sensitivity analysis on $\gamma$ in structure simplicity ($S$).\looseness=-1}
\caption{The triple/fan accuracy in the sensitivity analysis on $\gamma$ in structure simplicity ($S$) with different knowledge bases.\looseness=-1}
\label{table:sensitivity}
\setlength{\tabcolsep}{1.6mm}{
\begin{tabular}{ccccccc}
\hline
$\gamma$ & 0 & 1 & 2 & 3 & 4 & 5 \\
\hline
DBpedia & 0.321 & 0.351&0.496&0.542&0.619&0.589 \\
AMiner SciKG & 0.633&0.638&0.639&0.635&0.633&0.627 \\
\hline
\end{tabular}
}
\end{table}

\subsubsection{Performance of Constrained Clustering}

In this experiment, we evaluated the quality of the \weikai{clustering tree} by using three real-world datasets.

\begin{figure}[!tb]
\centering
\subfigure[New York Times dataset]{\includegraphics[width=\linewidth]{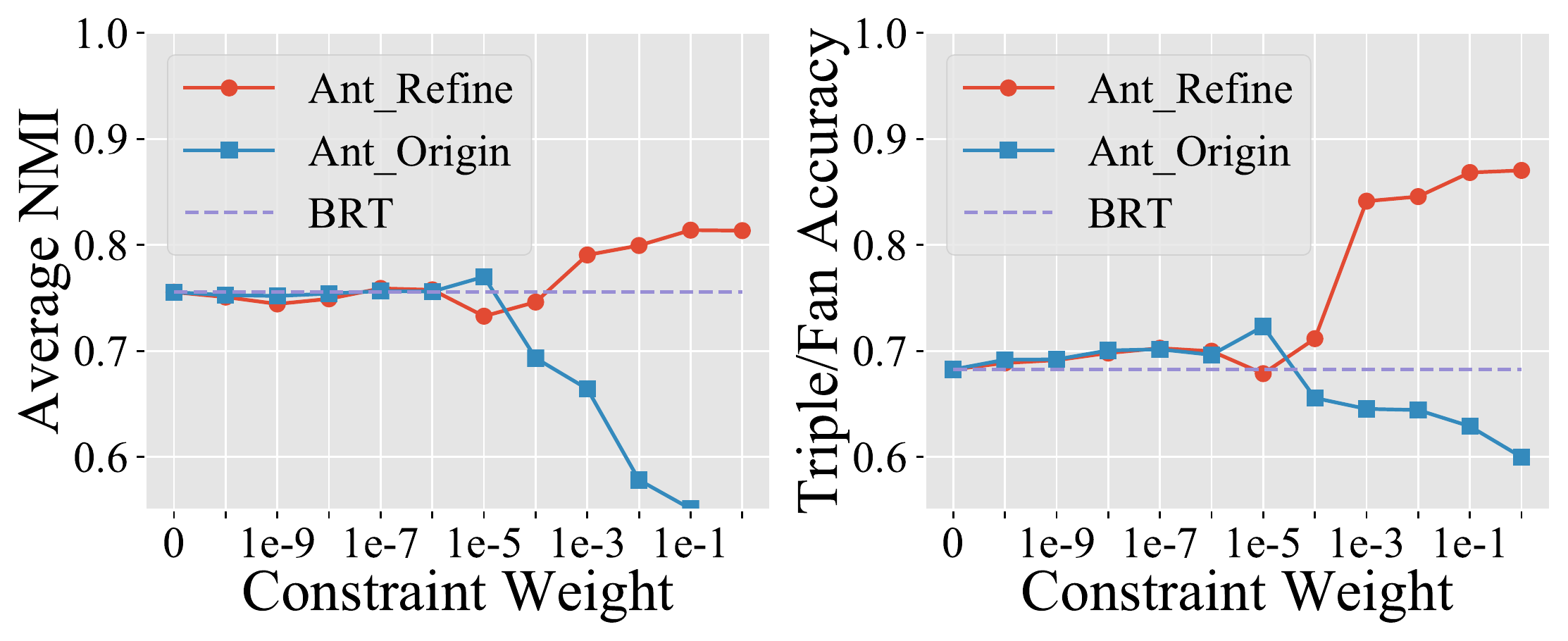}}
\subfigure[20 Newsgroups dataset]{\includegraphics[width=\linewidth]{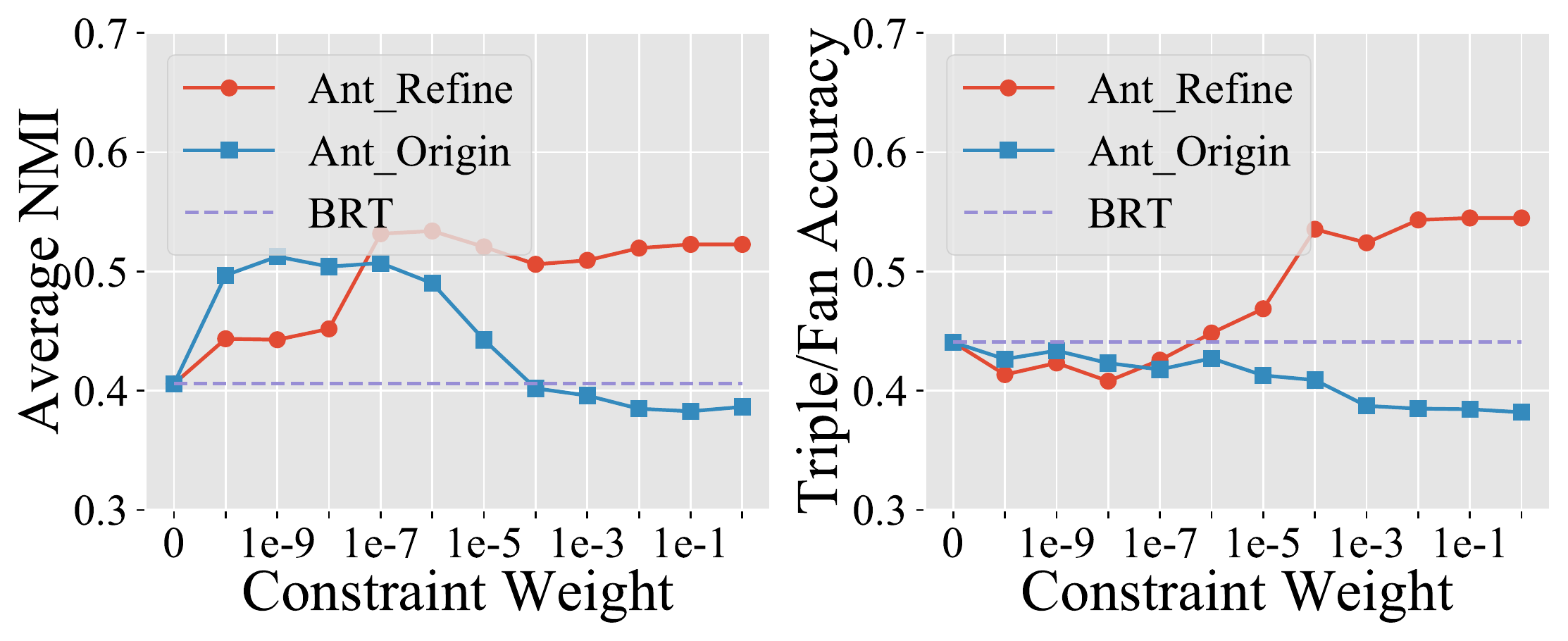}}
\subfigure[AMiner dataset]{\includegraphics[width=\linewidth]{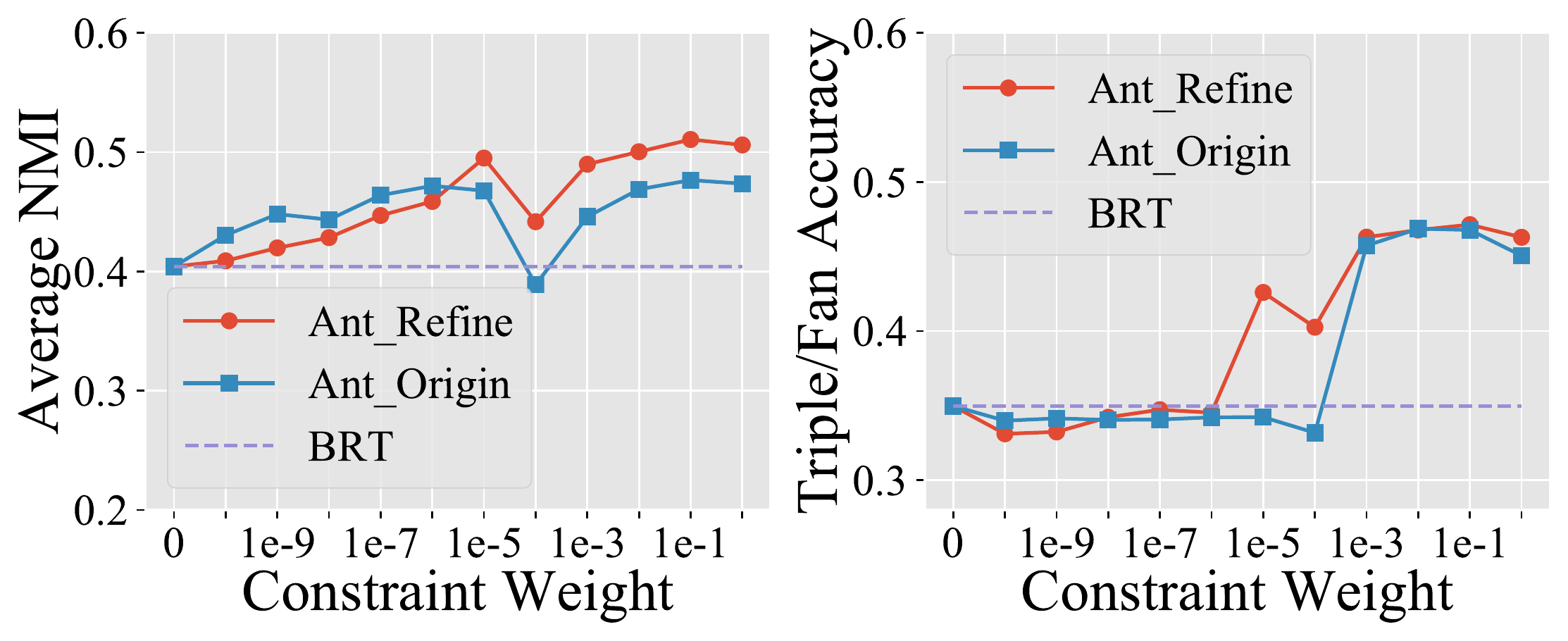}}
\caption{Comparison of hierarchical clustering quality.
}
\label{fig:clutesring_evaluation}
\end{figure}

\noindent \textbf{Evaluation criteria.}
We evaluated the hierarchical clustering results by comparing the \weikai{clustering trees} with the ground-truth hierarchies provided in the datasets.
In addition to triple/fan accuracy, we also measured the \textbf{average NMI} (Normalized Mutual Information).
NMI is widely-used to assess the similarity between two clustering results~\cite{strehl2002cluster}.
However, it is designed for non-hierarchical scenarios.
To assess hierarchical clustering quality, we computed NMI scores for each layer and averaged the scores.
Larger average NMI scores indicate higher quality.
\weikaiminor{Careful considerations went into selecting the most appropriate metric for evaluating the performance of constrained clustering. We chose not to use unsupervised clustering metrics such as Silhouette coefficient~\cite{rousseeuw1987silhouettes} as they are based only on instance similarity and do not consider any knowledge used to influence the clustering result.}

\noindent \textbf{Results.}
Fig.~\ref{fig:clutesring_evaluation} compares our method with BRT in terms of hierarchical clustering quality on three real-world datasets.

\emph{Overall performance}.
We observed that Ant\_Origin performed better than BRT in terms of average NMI and triple/fan accuracy, and the result was further improved in Ant\_Refine.
This shows that better constraints lead to better clustering results,
which demonstrates the effectiveness of our constrained clustering method.

\emph{Effect of constraint weights}.
\weikai{The constraint weights indicate to what extent the constraints will affect the clustering results.}
Since BRT does not consider constraints, its results are the same for different constraint weights.
For our method, the clustering quality usually increases with increasing constraint weight initially.
When the weight becomes larger (e.g., 1e-3), the clustering quality decreases with increasing constraint weight.
Our method achieved the best performance when the constraint weight is between 1e-7 and 1e-5. 
This demonstrates that both the knowledge-driven term (constraints) and the data-driven term are important.
Ignoring either one of them usually degrades the performance. 

\weikai{
\subsubsection{Efficiency of Constrained Clustering}\label{subsec:efficiency}
\xitingrevision{In this experiment, we evaluated the time cost for extracting constraint trees and perform constrained clustering.}
}

\noindent\textbf{Experimental settings.}
\weikai{The knowledge base used in the experiments was DBpedia, which contains \wenwen{approximately} 885,000 categories. 
We sampled $n$ documents from \textit{category set B} in the 20 Newsgroups dataset ($n\in \{1000,2000,\ldots,5000\}$).
The experiments were conducted on a desktop PC with an Intel i7-9700k CPU (3.6 GHz) and 32 GB RAM.
To eliminate the randomness caused by the path selection in the ant colony optimization algorithm, we repeated each experiment 10 times. The results reported are the average values of the 10 trials.
}

\noindent\textbf{Results.}
As shown in Table~\ref{table:efficiency}, the time required for constraint extraction is roughly linear to the number of documents.
It is reasonable since both the projection step and ant colony optimization step take $O(n)$ time ($n$ is the number of documents).
According to the experiment, it takes nearly 3 minutes to extract the relevant constraints for a corpus with 5,000 documents based on a very large knowledge base, DBpedia. 
As constraint extraction is an offline process, such a time cost is acceptable. 
The constrained clustering method has a higher time complexity of $O(n^2)$, but it is still capable of hierarchically clustering documents efficiently when the number of documents is not very large.
For example, it can handle 5,000 documents within one minute.
\begin{table}[htb]
\caption{Time cost comparison (in seconds) of constraint extraction and constrained clustering on different numbers of documents.\looseness=-1}
\label{table:efficiency}
\setlength{\tabcolsep}{1.8mm}{
\begin{tabular}{cccccc}
\hline
Number of documents & 1000 & 2000 & 3000 & 4000 & 5000 \\
\hline
Constraint extraction & 42.0 & 74.8 & 108.1 & 140.3 & 172.0 \\
Constrained clustering & 4.1 & 12.6 & 24.2 & 38.6 & 58.7 \\
\hline
\end{tabular}
}
\end{table}

\subsection{Case Study}
In this section, we demonstrate how a domain expert leveraged ReVision to quickly organize literature to benefit his research.

E1 is a Microsoft research scientist whose job duty involves developing interactive machine learning techniques.
He would like to follow the recent developments in machine learning research.
As a start, E1 selected a few seed publications about interactive machine learning and expanded the selection by adding the papers referenced by the seed ones.
He then treated the AMiner Science Knowledge Graph~\cite{data_aminer} as the public knowledge source, which revealed the relationships between the concepts in the domain of computer science.
Leveraging this knowledge base, he created a constraint tree in ReVision (\textbf{R1}).
With the initial constraints, E1 was able to focus on a coarse-grained structure adjustment.

\noindent\textbf{Examine the initial constraints (R2.1, R2.2).}
As shown in Fig.~\ref{fig:teaser-overview}(b), the extracted constraints consisted of four branches, ``Information System,'' ``Human-centered Computing,'' ``Computing Methodologies,'' and ``Mathematics of Computing.''
Labels from the knowledge base helped E1 understand the content and structure of the constraint hierarchy. 
For instance, he quickly located a few important nodes such as ``Natural Language Processing (NLP),'' ``Machine Learning (ML),'' and ``Computer Vision (CV)'' under the branch ``Computing Methodologies'' (Fig.~\ref{fig:teaser-overview}(F)).
He also found ``Visual Analytics'' in the same branch, which should not be placed here in his opinion (\textbf{R2.1}).
The grain glyph drew his attention to the uncertain parts of the constraints (\textbf{R2.2}).
For example, he noticed that ``World Wide Web'' under ``Information System'' and ``Discrete Mathematics'' under ``Mathematics of Computing'' had higher uncertainty (Fig.~\ref{fig:teaser-overview}(A)(B)).\looseness=-1

\noindent\textbf{\weikai{Node}-level refinement of the constraints (R2.3).}
After a quick examination of the initial constraints, E1 began to refine the initial constraints through \weikai{node}-level refinement.

First, based on his preference in organizing the relevant research areas, E1 moved the branch ``Visual Analytics'' to ``Human-centered Computing.'' 
He then reorganized the nodes under ``Computing Methodologies'' based on their relevance to his research interest by moving ``NLP'' and ``CV'' to the same level as ``Machine Learning'' (Fig.~\ref{fig:teaser-overview}(F)(G)).

Next, he clicked ``World Wide Web'' to investigate the reason for its high uncertainty.
The word cloud in the information panel revealed high-frequency words, such as ``twitter'' and ``user'' (Fig.~\ref{fig:teaser-overview}(D)).
The titles of the papers showed that they were about visual analysis research leveraging the social network for tasks such as personalization and event detection (Fig.~\ref{fig:teaser-overview}(E)).
Hence he renamed the node to ``Social Media Analysis'' and moved it to ``Visual Analytics.''

Finally, he focused on another highly uncertain branch, ``Discrete Mathematics,'' which seemed irrelevant to machine learning or visualization.
After checking the titles and the high-frequency words, he found that the papers were about visualization techniques for trees or directed graphs and were related to ``Graph Theory'' (a child of ``Discrete Mathematics''). 
Another child, ``Combinatorics,'' contained only a few documents, and they were far away from his interest.
Therefore, he removed ``Combinatorics'' and dragged ``Graph Theory'' to join node ``Visualization Techniques'' under ``Visualization.''

After the aforementioned modifications, E1 was satisfied with the constraints.
He then built the \weikai{clustering tree} based on the refined constraints and examined it.

\begin{figure}[!b]
\centering
\includegraphics[width=\linewidth]{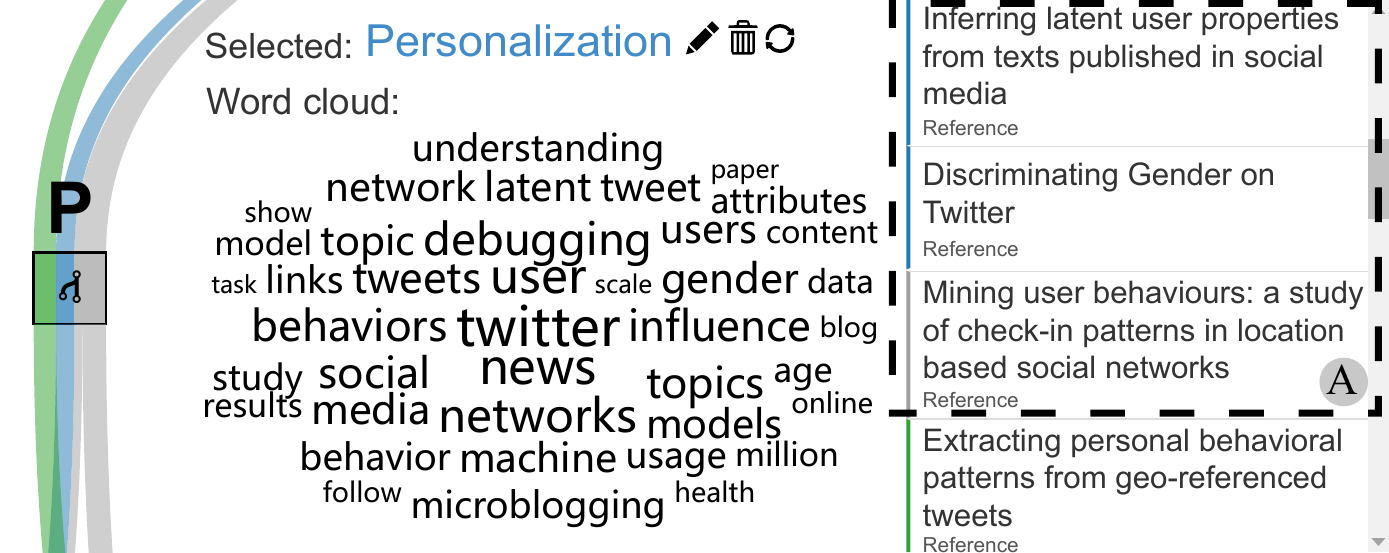}
\caption{A node labeled ``Personalization'' in the \weikai{clustering tree}. A few papers belonging to ``Computing Methodologies'' (blue) were mixed in.}
\label{fig:case_person}
\end{figure}

\noindent\textbf{Document-level refinement (R2.3).}
E1 further fine-tuned the hierarchy by performing document-level adjustments.
In the \weikai{clustering tree}, he found a node labeled ``Personalization,'' where the documents were about the analysis on social media.
However, a few papers belonging to ``Computing Methodologies'' (blue) were mixed in (Fig.~\ref{fig:case_person}(A)).
To figure out the potential reason for such a mixture, he switched to the constraint tree and examined the document distribution of this node. 
He found that most documents were under ``Social Media Analysis'' (green).
However, a few documents, such as ``Inferring Latent User Properties from Texts Published in Social Media,'' belonged to ``NLP'' under ``Computing Methodologies'' (blue) in the constraint tree.
After a careful examination, he moved these documents from ``NLP'' to ``Social Media Analysis.''

\noindent\textbf{Cluster newly published papers incrementally (R2).}
E1 was interested in the recent research trends in NLP and machine learning, so he collected the relevant papers from the ACL and NeurIPS conferences and treated the refined constraint tree as a knowledge base for the constrained clustering method. 

\begin{figure}[!tb]
\centering
\includegraphics[width=\linewidth]{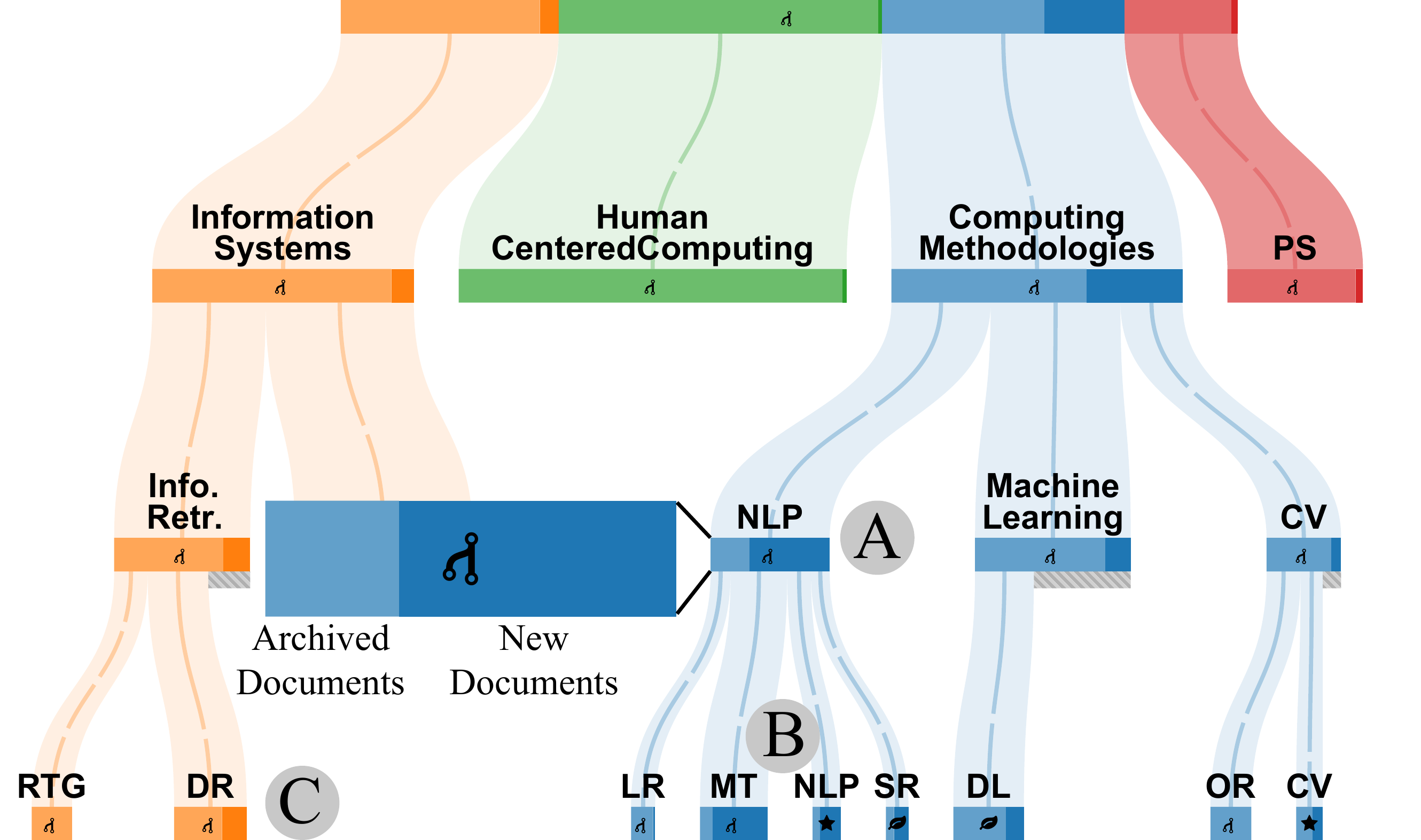}
\caption{Most of the documents were automatically added to ``NLP,'' while some of them were added to ``Document Representation (DR).'' The proportion of the darker bar in the node represents the proportion of the new documents.}
\label{fig:case_add_acl}
\end{figure}

He first added the papers from ACL 2017 and 2018.
As expected, most ACL papers were classified into node ``NLP'' in the constraint tree (Fig.~\ref{fig:case_add_acl}(A)).
He found that the substructure (e.g., ``Machine Translation'' and ``Speech Recognition'') provided by the constraint tree successfully organized most papers (Fig.~\ref{fig:case_add_acl}(B)). He also noticed that ``Machine Translation'' captured more new papers than ``Speech Recognition,'' indicating that the former has become a more popular research area in recent years (\textbf{R2.1}).
A further examination showed that a few papers went into node ``Information System -- Information Retrieval -- Document Representation (DR)'' (Fig.~\ref{fig:case_add_acl}(C)).
Since ``Document Representation'' contained many papers about the topic model, one of the major research areas of NLP, he preferred to move it into ``NLP'' (\textbf{R2.3}).

\begin{figure}[!b]
\centering
\includegraphics[width=\linewidth]{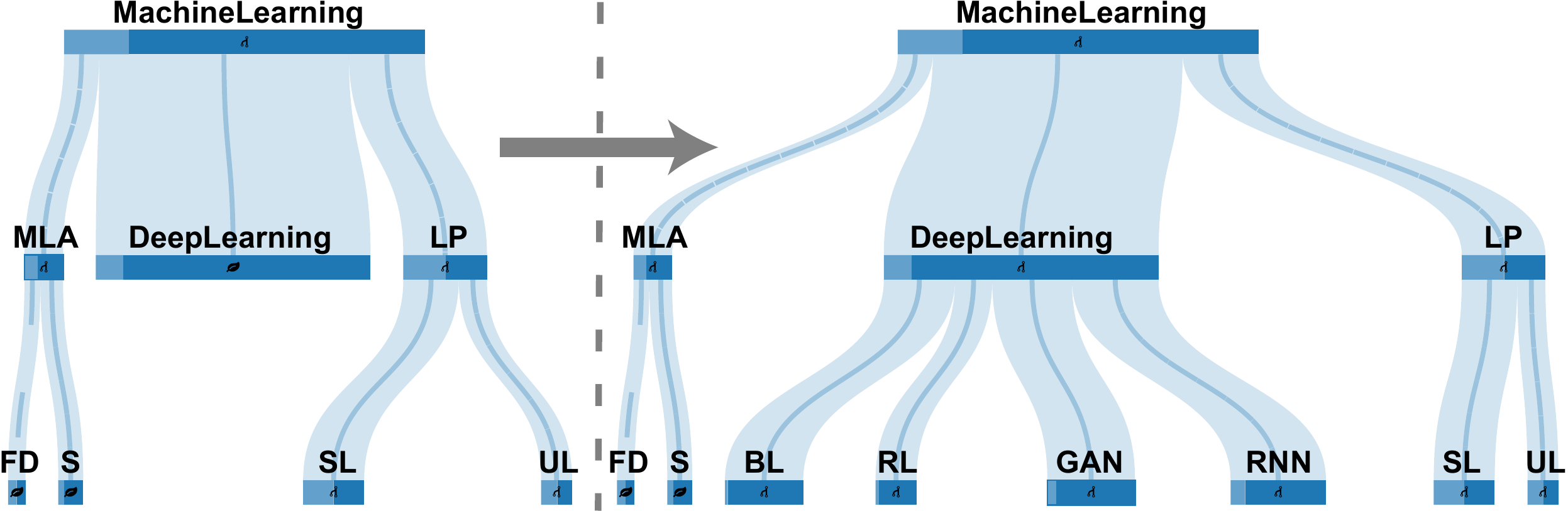}
\caption{E1 rebuilt the hierarchy for the large leaf node ``Deep Learning'' and found more interesting topics.}
\label{fig:case_divided_dl}
\end{figure}

E1 added papers from the NeurIPS 2017 and 2018.
Some papers were automatically assigned to the nodes (e.g., papers about matrix factorization were added to ``nonnegative matrix factorization'') in the constraint tree.
However, when he expanded ``Computing Methodologies,'' which contained many new papers, he found the constraint tree could not keep up with the quick development of AI-related research.
For example, ``Deep Learning'' was a small leaf node in the initial constraint tree. 
As more papers were published on deep learning, the initial constraint tree failed to provide a detailed structure for this area (\textbf{R2.1}).
Therefore, he rebuilt the hierarchy by applying hierarchical clustering on these documents directly and identified several important nodes such as ``Bayesian Learning,'' ``Reinforcement Learning,'' ``Generative Adversarial Network,'' and ``Recurrent Neural Network'' (\textbf{R2.3}) (Fig.~\ref{fig:case_divided_dl}).
He applied the refined constraints to the clustering process and obtained a new clustering tree, which helped him find more interesting topics.\looseness=-1

%% file: discussion.tex
\section{Discussion and Future Work}
\label{sec:discussion}

\weikai{
\noindent \textbf{Latency in visualization.}
In the visualization, the adjustment on the constraints does not immediately trigger the update of the model, so these operations can finish within several hundreds of milliseconds and cause little latency.
However, it may take several seconds to rebuild the clustering result after users finish the refinement and click ``Update,'' or when clicking ``Add Papers'' to add more data.
We have discussed the efficiency in \xitingrevision{Sec.}~\ref{subsec:efficiency}.
Typically, it takes less than one minute to update the clustering result for 5,000 documents.\looseness=-1
}

\noindent \xitingrevision{\textbf{\normalsize Extension to Other Data Types.}}
In this paper, we use textual data as a guiding example to illustrate how the interactive steering of hierarchical clustering can be achieved. % by using our method
Our method can easily be adapted to other types of data (e.g., images or users in a social network).
\xitingrevision{
In our method, only the calculation of $f_{\text{DCM}}(d,v)$ depends on the type of data.
Leveraged in Eq.~(\ref{eq:accuracy}) for path accuracy estimation, $f_{\text{DCM}}(d,v)$ measures the probability that data item $d$ belongs to cluster $v$.}
For textual documents, DCM distribution is an appropriate \xitingrevision{probability distribution}.
For other types of data, we can replace DCM with other distributions.
\xitingrevision{For example, we can cluster images or users in a social network by normalizing the image or user feature vectors and replacing $f_{\text{DCM}}(d,v)$ with the Von Mises–-Fisher distribution $f_{\text{VMF}}(d,v)$~\cite{jupp1979maximum}.}
In the future, we plan to apply our method to handle other types of data.
Another interesting aspect to investigate is how to cluster heterogeneous data items (data items of multiple types). %\looseness=-1

\noindent \textbf{Intelligent Refinement Suggestion.}
Currently, we provide visual cues to highlight the structures that have high uncertainties.
We may further reduce user efforts by providing suggestions on how we can refine the hierarchy.
For example, after a user drags cluster ``Rocket'' from ``Transport'' to ``Space,'' 
we may suggest that clusters ``Spaceflight'' and ``Spacecraft'' can be moved to ``Space.''

\noindent \textbf{Collaborative Clustering.}
Our method provides a natural way for building hierarchies collaboratively: hierarchies constructed by one user can be considered as constraint trees or public knowledge for other users.
In the future, we are interested in identifying important real-world applications for building hierarchies with a group of people and evaluate how effective our method is in these scenarios.

%% file: conclusion.tex
\section{Conclusion}\label{sec:conclusion}
In this paper, we present ReVision, an interactive visual analysis system to interactively steer constrained hierarchical clustering by leveraging knowledge from the public domain and individual users. 
To better support the steering of clustering, our approach first constructs constraints for otherwise unsupervised hierarchical clustering to improve the initial results. 
Users can then incorporate their own knowledge in an efficient way to steer the clustering algorithm through an interactive visual interface that clearly presents uncertainty information about the clustering results. 
A thorough quantitative evaluation is performed on all computational components of ReVision and demonstrated the advantages of the chosen methods.
In addition, the case study showcases how ReVision enables users to refine the hierarchy quickly and to meet their customized needs.